\documentclass[runningheads]{llncs}

\usepackage{graphicx}
\usepackage{comment}
\usepackage{amsmath,amssymb}
\usepackage{color}
\usepackage{url}
\usepackage{hyperref}

    \usepackage{wrapfig}
    \usepackage[percent]{overpic}
    \usepackage[dvipsnames,table]{xcolor}
    \usepackage{adjustbox}
    \usepackage{enumitem}
    \usepackage{rotating}
    \usepackage{subcaption}
    \usepackage{graphicx}
    \usepackage{multirow}

    \setlist[itemize]{noitemsep,leftmargin=*,topsep=0em}
    \setlist[enumerate]{noitemsep,leftmargin=*,topsep=0em}
    \usepackage[capitalize]{cleveref}
    \crefname{section}{Sec.}{Secs.}
    \Crefname{section}{Section}{Sections}
    \Crefname{table}{Table}{Tables}
    \crefname{table}{Tab.}{Tabs.}

\newif\ifdraft
\draftfalse
\drafttrue

\definecolor{orange}{rgb}{1,0.5,0}
\definecolor{violet}{RGB}{70,0,170}
\definecolor{marcs_green}{rgb}{0.26, 0.78, 0.40} 

\definecolor{plot_blue}{RGB}{67,138,189}
\definecolor{plot_orange}{RGB}{243,151,58}

\definecolor{hist_cars}{RGB}{166,197,221}
\definecolor{hist_chairs}{RGB}{248,201,158}
\definecolor{hist_planes}{RGB}{174,212,170}

\iffalse
 \newcommand{\PF}[1]{{\color{blue}{\bf PF: #1}}}
 
 \newcommand{\BG}[1]{{\color{red}{\bf BG: #1}}}
 \newcommand{\bg}[1]{{\color{red} #1}}
 \newcommand{\MH}[1]{{\color{marcs_green}{\bf MH: #1}}}
 
 \newcommand{\CT}[1]{{\color{olive}{\bf LD: #1}}}
 
\else
 \newcommand{\PF}[1]{}
 
 \newcommand{\BG}[1]{}
 \newcommand{\bg}[1]{#1}
 \newcommand{\MH}[1]{}
 
 \newcommand{\CT}[1]{}
 
\fi

\newcommand{\baselineRelu}{\textit{ReLU-net}}
\newcommand{\baselineSin}{\textit{Sin-net}}

\newcommand{\bx}{\mathbf{x}}
\newcommand{\bl}{\mathbf{l}}
\newcommand{\bL}{\mathbf{L}}
\newcommand{\bz}{\mathbf{z}}
\newcommand{\bW}{\mathbf{W}}
\newcommand{\bb}{\mathbf{b}}

\newcommand{\BR}{\mathbb{R}}
\newcommand{\bbg}{\mathbf{g}}
\newcommand{\bomega}{\boldsymbol{\omega}}
\newcommand{\bphi}{\boldsymbol{\phi}}

\newif\ifreview
\reviewfalse

\ifreview
	\usepackage{lineno}

	\linenumbers
\fi

\begin{document}


\def\SubNumber{23}

\def\GCPRTrack{Main Track}

\title{A Latent Implicit 3D Shape Model for Multiple Levels of Detail}

\ifreview
\else

	\author{Benoit Guillard\inst{1} \and
	Marc Habermann\inst{2} \and
	Christian Theobalt\inst{2} \and
        Pascal Fua\inst{1}} 
	
	\authorrunning{B. Guillard et al.}
	
	\institute{CVLab, EPFL, Switzerland \email{\{benoit.guillard, pascal.fua\}@epfl.ch} \and Max Planck Institute for Informatics, Saarland Informatics Campus, Germany
	\email{\{mhaberma, theobalt\}@mpi-inf.mpg.de}}
\fi

\maketitle              



\begin{abstract}
Implicit neural representations map a shape-specific latent code and a 3D coordinate to its corresponding signed distance (SDF) value.
However, this approach only offers a single level of detail.
Emulating low levels of detail can be achieved with shallow networks, but the generated shapes are typically not smooth.
Alternatively, some network designs offer multiple levels of detail, but are limited to overfitting a single object.

To address this, we propose a new shape modeling approach, which enables multiple levels of detail and guarantees a smooth surface at each level.
At the core, we introduce a novel latent conditioning for a multiscale and bandwith-limited neural architecture.
This results in a deep parameterization of multiple shapes, where early layers quickly output approximated SDF values.
This allows to balance speed and accuracy within a single network and enhance the efficiency of implicit scene rendering.
We demonstrate that by limiting the bandwidth of the network, we can maintain smooth surfaces across all levels of detail.
At finer levels, reconstruction quality is on par with the state of the art models, which are limited to a single level of detail.

\end{abstract}



\begin{center}
        \begin{overpic}[width=0.9\textwidth]{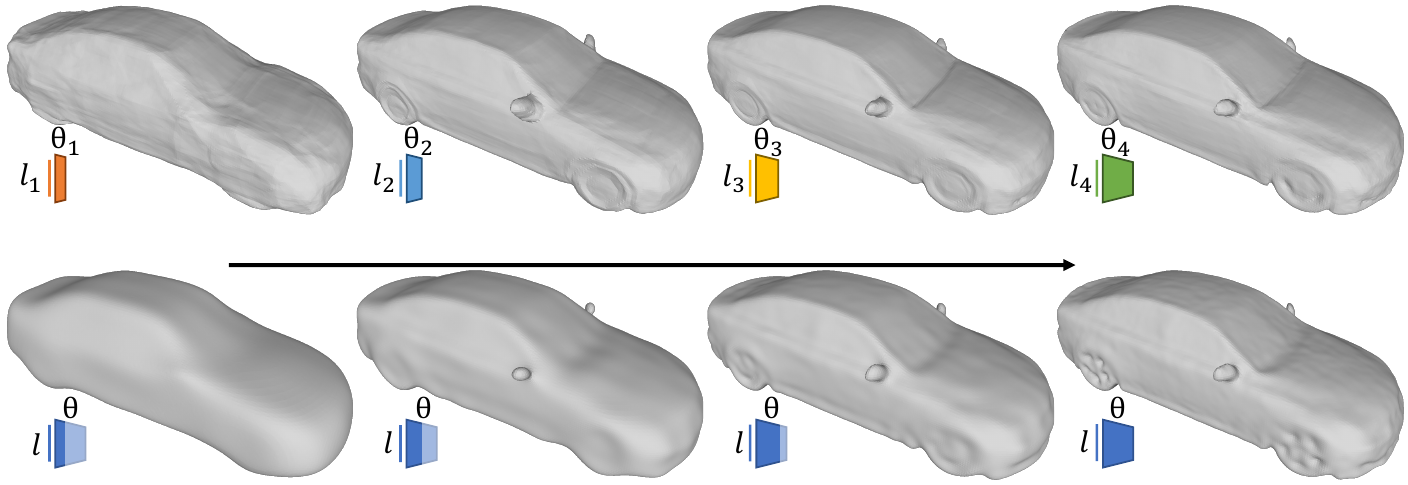}
            \put(13,16.5){Increasing level of detail and computation time}
            \put(-2,21){\begin{turn}{90} \small ReLU net.\end{turn}}
            \put(-2,1){\begin{turn}{90}\textbf{ \small Our net.}\end{turn}}
        \end{overpic} \\
\end{center}
\vspace{-5mm}

\noindent  \hypertarget{fig:teaser}{\begin{small}Implicit neural shape representations map latent codes to 3D surfaces a using a neural network.
\textbf{Top:} traditional networks can only reconstruct shapes at one level of detail (LoD). To model increasing LoDs, networks must be made deeper ($\theta_1$ to $\theta_4$). These networks do not share their latent codes $l_i$, and yield rough surfaces when shallow ($i=1$).
\textbf{Bottom:} we propose to use a single network $\theta$ to map the same latent code $l$ to multiple LoDs. Lower LoDs are quickly provided by the early layers of the network. We restrict their bandwidths with a new network design to reconstruct smoother shape approximations at low LoD.
\end{small}} 



\section{Introduction}
Latent implicit shape representations~\cite{Park20a,Mescheder19,Jiang20,Chabra20,Li22b} have become increasingly important for 3D shape modeling. 
They learn a parametric representation of 3D shapes in the form of latent vectors.
They streamline the modeling process, by reducing it to finding a single latent vector. However, they only provide a single level of detail (LoD) and sometimes produce unnecessarily bumpy surfaces when using shallow network architectures. 
This is regrettable because LoD control
allows to balance computational speed and surface precision.
For instance, in 3D scene rendering, objects that are far away from the virtual camera do not require high geometric details. 
Similarly, smooth surfaces benefit many downstream applications as they allow for more natural shading and better rendering.
\par
The most trivial way to represent implicit shapes at multiple LoD is to simply use shallower networks. 
This requires one network per level of detail, each of them with different latent spaces.
Instead, our goal is that multiple LoD share the same latent space.
Therefore, we start from an existing multiscale and band-limited architecture~\cite{Lindell22} that ensures that geometric details at certain layers do not exceed a given frequency, but is designed to model a single signal. 
We show how it can be extended to provide a latent representation for 3D surfaces while preserving its band-limited properties, which ensures smooth surfaces at all LoD.
With our design, early layers quickly provide reasonable shape estimates and later layers simply add more details, which makes it easy to select the desired level of detail for any potential application. 
We explain how this can be exploited to quickly explore the shape space in an interactive manner.
\par 
In summary, our contributions are as follows:
\begin{itemize}
    \item We propose the first latent and implicit 
     method, which supports multiple level of details and guarantees surface smoothness.
    \item To this end, we introduce latent conditioning to a multiscale and band-limited neural network architecture, while preserving its band-limiting property.
    \item We also introduce a joint optimization of the frequency and phase parameters of each positional encoding layer, resulting in higher quality shape models.
    \item We demonstrate efficient latent space exploration for interactive shape editing and design.
\end{itemize} 
Our experiments demonstrate that the proposed method significantly improves the smoothness across all LoD of the surface while demonstrating a geometric quality that is on par with the current state of the art implicit shape modeling methods, which do not support multiple levels of detail. 
We will release code upon acceptance.
%


\section{Related Work}
\textbf{Implicit neural shape representations} were first introduced in DeepSDF~\cite{Park20a} and Occupancy networks~\cite{Mescheder19}.
These representations use a fully connected network to map a latent code to a 3D scalar field, which is converted to a watertight shape by extracting its isosurface with Marching Cubes~\cite{Lewiner03}.
They were originally only suitable for watertight surfaces, but later extended to unsigned distance fields \cite{Chibane20b,Zhao21a,Guillard22b}. 
These representations have been applied in various tasks, such as single view reconstruction of objects \cite{Xu19b} and humans \cite{Saito19a,Saito20a} from images or point clouds \cite{Peng20c}.
They were also improved with regards to various aspects, such as the smoothness of the latent space \cite{Liu22} or the physical plausibility of the output shapes \cite{Mezghanni22}

Furthermore, part-wise representations have been obtained by splitting the latent code in 3D space \cite{Jiang20,Chabra20,Li22b}, and transformer layers have been applied to refine these part-wise 3D latent codes \cite{Hertz22,Zhang22}.
Despite providing very accurate reconstructions, they only provide one level of detail, and no longer provide shape parameterizations from a single vector.
In comparison, our network relies on a compact latent code, which enables shape recovery and manipulation in the presence of an incomplete signal \cite{Remelli20b}.
\par 
\textbf{Multiple levels of detail} \bg{for 3D shapes are traditionally obtained by post-processing meshes~\cite{hoppe2023progressive,hoppe1997view,wang2015rolling,zhang2018static}.
By contrast, to save computations, we aim at achieving variable LoD from the implicit network itself, before meshing the implicit signal.
Variable LoD}
have already been incorporated into implicit networks, but none of them learn a latent shape model.
One approach is to store latent codes at multiple levels of an octree, which are then decoded with a shared network to provide multiple levels of detail, as in NGLOD \cite{Takikawa21}.
Another approach is \textsc{Bacon}~\cite{Lindell22}, which stores implicit signals at multiple levels of detail within the weights of a single network.
It is using a specific band-limited architecture \cite{Fathony21} to \bg{progressively} improve coarse approximations \cite{Lindell22} \bg{with each layer}.
The residual formulation of \cite{Shekarforoush22} improves its performance on hierarchical optimizations problems, and it has been extended in \cite{Yang22} to impose lower bounds on bandwidth and specific phases \bg{for the signal}.
SAPE \cite{Hertz21} also learns variable frequency encodings, \bg{and \cite{takikawa2022variable} demonstrates high compression rates of neural radiance fields using vector quantization.}
However, as \textsc{Bacon}, they remain limited to overfitting a single signal.
Some studies have attempted to guide the training of implicit networks by gradually increasing the level of detail during training \cite{Duan20}, but they remain limited to one level at inference.
Closer to our work is MDIF \cite{Chen21}, which learns a multiscale representation on a grid of codes and uses a drop-out strategy for enabling shape completion. This requires writing per-task regularization losses when full volumetric supervision is not available for unseen shapes. By contrast, our representation is more versatile since it relies on a single latent code.

\textbf{New architectures} departing from standard fully connected netwoks have been proposed for implicit networks, to tailor them to the task at hand.
For instance, activation functions can be replaced by Gaussian in implicit networks \cite{Ramasinghe22}, and SIREN \cite{Sitzmann20} uses periodic activations in conjunction with a hypernetwork to learn a shape prior. 
Dupont et al.~\cite{Dupont22} take a different approach by using a SIREN network modulated by latent codes to learn shape priors, which enables a single network to represent multiple shapes. 
SIREN is also used in \cite{Yifan22}, which combines two networks to learn an SDF from a point cloud input, representing a coarse shape and higher frequency displacements. 
Our work also departs from standard fully connected networks, with the aim of jointly achieving conditioning on a latent code and imposing band-limiting at multiple scales.


%
\section{Method}
Our goal is to learn an implicit shape model from a collection of 3D shapes of a specific category, e.g. cars or chairs, which supports \textit{multiple levels of detail} and at each level \textit{smoothness} is also guaranteed. 
We learn a function 
%
\begin{equation}
    f_\Omega(\mathbf{x}, \mathbf{l},i) = s
\end{equation}
%
predicting the signed distance value $s \in \mathbb{R}$, where $\mathbf{x} \in \mathbb{R}^3$ is a query point in 3D space, $\mathbf{l} \in \mathbb{R}^{d_l}$ is a latent code for modeling the shape, $i \in \{1,..., N-1\}$ is the desired level of detail, and $\Omega$ are the learnable parameters of the network.
Inspired by \textsc{Bacon}~\cite{Lindell22}, we first recap how a band-limited network architecture can implement a simplified version of the above function, i.e. $f_\Omega(\mathbf{x},i)$, where latent conditioning and therefore modeling different shapes is not possible (\cref{sec:bacon}).
Next, we discuss multiple possible solutions for introducing a latent conditioning to this type of architecture and provide an analysis why only one of the proposed designs can accomplish this goal while preserving the original properties of band-limited networks
(\cref{sec:latent_conditioning}).
To achieve a higher shape accuracy, we further introduce a joint learning of per-layer frequencies and phases (\cref{sec:learn_frequency}).
By exploiting the multiple levels of detail, we then introduce an interactive latent space exploration enabling efficient shape manipulation (\cref{sec:exploration}).
Last, we provide details for the training and inference procedures (\cref{sec:training}).
%
%
\subsection{Band-limited Coordinate Network} \label{sec:bacon}
A \textsc{Bacon} network~\cite{Lindell22} with $N$ layers maps an input coordinate $\bx \in \BR^3$ to $N - 1$ estimates of the SDF at this position, $f_{\Omega}(\bx, i)$ with $1 \leq i < N$. 
The input coordinate $\bx \in \BR^3$ is first encoded using sine layers 
%
\begin{align}
\label{eq:coord_embedding}
    \bbg_i(\bx)=\sin(\bomega_i \bx + \bphi_i) \in \BR^{d_h}
\end{align}
%
for $i=0,...,N-1$, with \bg{\textit{frequencies}} $\bomega_i \in \BR^{d_h \times 3}$, \bg{\textit{phase shifts}} $\bphi_i \in \BR^{d_h}$, and $d_h$ the hidden dimension.
Intermediate activations $\bz_i \in \BR^{d_h}$ and SDF outputs $f_{\Omega}(\bx,i) \in \BR$ are
%
\begin{align}
\label{eq:bacon}
    \bz_0 &= \bbg_0(\bx) \mbox{ and } \bz_i = \bbg_i(\bx) \circ (\bW_i \bz_{i-1} + \bb_i) \; , \: \:  1 \leq i < N \nonumber \\ 
    f_{\Omega}(\bx,i) &= \bW_i^{out} \bz_i + \bb_i^{out} 
\end{align}
%
where $\circ$ is the Hadamard product, ${(\bW_i, \bb_i) \in (\BR^{d_h \times d_h} \times \BR^{d_h})}$ are the \textit{linear layers} and ${(\bW_i^{out}, \bb_i^{out}) \in (\BR^{1 \times d_h} \times \BR)}$ are the \textit{output layers}.
The learnable parameters of the network ${\Omega = \{\bW_i, \bb_i, \bW_i^{out}, \bb_i^{out} \}}$ are overfit to a single SDF signal.
%
%
%
\bg{Frequency} coefficients $\bomega_i$ are sampled uniformly in $[-B_i, B_i]$, with $B_i > 0$, and in $[-\pi, \pi]$ for \bg{phases} $\bphi_i$,
%
\begin{align}
\label{eq:sampling_freq}
    \bomega_i \sim  \mathcal{U}([-B_i, B_i]^{d_h \times 3}) \; , \; \bphi_i \sim  \mathcal{U}([-\pi, \pi]^{d_h}) \; \; .
\end{align}
%
Applying the trigonometric property $2 \sin(a)\sin(b) = \sin(a+b-\tfrac{\pi}{2}) + \sin(a-b+\tfrac{\pi}{2})$
repeatedly to the Hadamard products of \cref{eq:bacon}, \cite{Fathony21,Lindell22} demonstrate that the output signal $f_{\Omega}(\cdot,i)$ of layer $i$ has its bandwidth upper-bounded by $\sum_{j=0}^i B_j$.
In other words, since multiplying sines sums their frequencies, this network design ensures that the maximum spatial frequency of the output SDFs is increasing with depth. 
As a result, 
(i) the first SDF outputs are smooth by construction, and 
(ii) the SDF fields $f_{\Omega}(\cdot,i)$ can represent more details as $i$ increases.
This increase is controlled by the bounds $B_i$ of \cref{eq:sampling_freq} and can be manually tuned.
%
%
\subsection{Conditional Band-limited Network} \label{sec:latent_conditioning}
%

\begin{figure*}[t]
    \centering
    \includegraphics[width=1\textwidth]{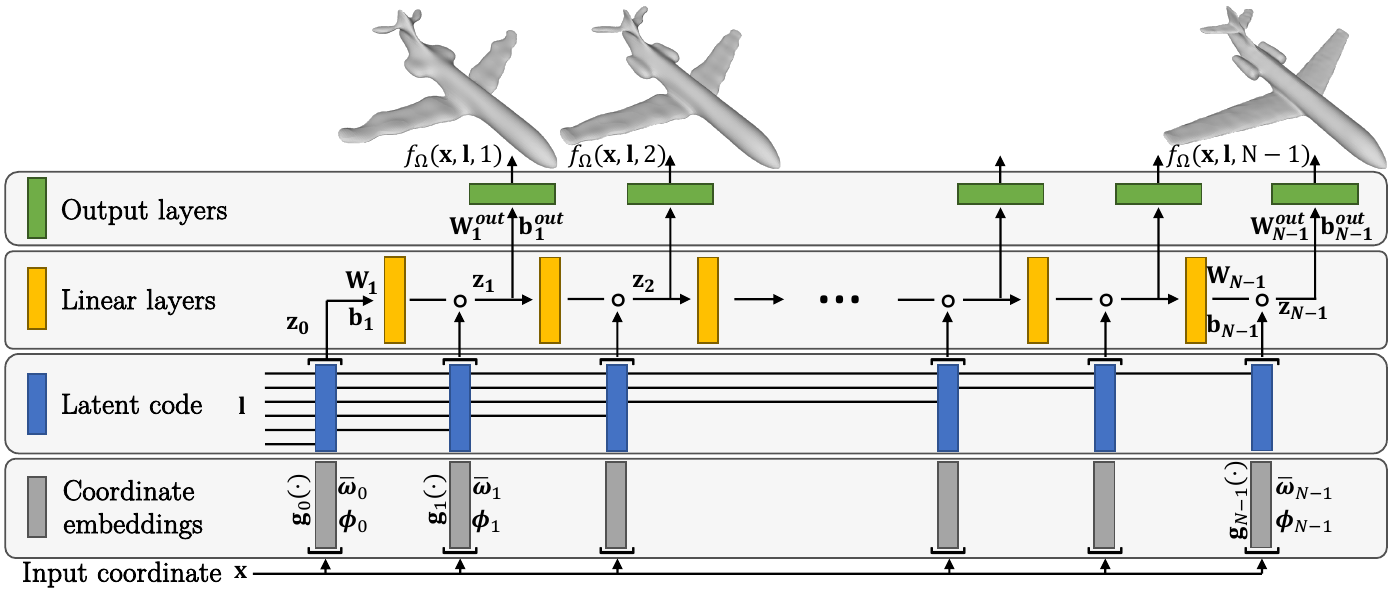}
    \vspace{-0.55cm}
    \caption{\textbf{Network Architecture.} 
    coordinate $\bx$ is encoded with parametric functions $\bbg_i(\cdot)$, and concatenated with the latent code $\bl$. 
    Linear layers $(\bW_i, \bb_i)$ and hadamard products yield the intermediate activations $\bz_i$. 
    At level $i$, the SDF output is predicted with the output layer $(\bW_i^{out}, \bb_i^{out})$.
    All network parameters ${\Omega = \{\bW_i, \bb_i, \bW_i^{out}, \bb_i^{out}, \overline{\bomega_i}, \bphi_i \}}$ are optimized jointly with the latent codebook.
    }
    \label{fig:pipeline}
    \vspace{-0.35cm}
\end{figure*}

\begin{wrapfigure}{R}{0.5\textwidth}
    \centering
    \vspace{-1cm}
    \begin{overpic}[width=0.48\textwidth,trim={0 0 0 -1.5cm},clip]{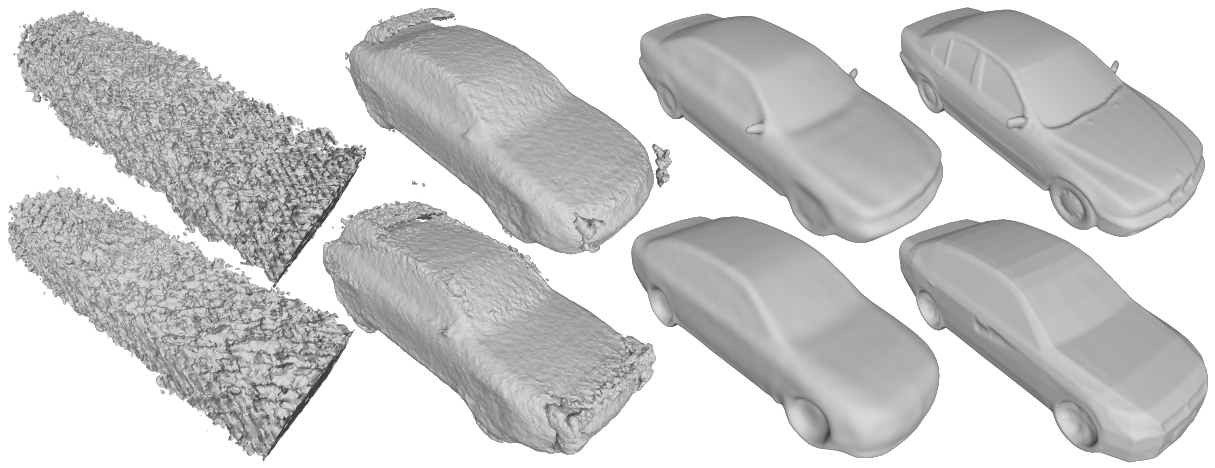}
            \put(0,38){{\small Design 1}}
            \put(26,38){{\small Design 2}}
            \put(50,38){{\small \textbf{Design 3}}}
            \put(82,38){{\small GT}}
    \end{overpic}
    \vspace{-0.25cm}
    \caption{
    \textbf{Architecture Design.} When concatenating a latent code to the input coordinates (\textbf{Design 1}) or the output layers (\textbf{Design 2}), the resulting shapes are unsatisfactory.
    When the latent code is concatenated to the hidden layers (\textbf{Design 3 (Ours)}), the generated shapes are significantly closer to the ground truth.
    %
    }
    \label{fig:design_exploration}
    \vspace{-0.45cm}
\end{wrapfigure}
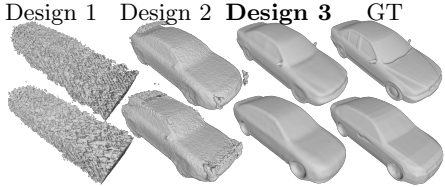

The network described above can only overfit a single SDF field, and, thus, represents a single shape as its $0$-level set. 
Our aim is to extend it to represent multiple geometries with a single network, by conditioning it on a shape specific latent code $\bl \in \BR^{d_l}$. 
Although the \textsc{Bacon} architecture effectively guarantees band-limited SDF outputs with respect to spatial coordinates, it is also a very constraining architecture that makes it challenging \bg{to incorporate a latent code}.
Consequently, in the remainder of this chapter, we explore and discuss three different alternatives for latent conditioning (see also \cref{fig:design_exploration}).
%
%
\par
\par \textbf{Design 1.}
The most trivial suggestion is to simply concatenate $\bl$ with the input coordinates $\bx$, i.e. \cref{eq:coord_embedding}, becomes
%
\begin{align}
    \bbg_i(\bx)=\sin(\bomega_i [\bx \;|\; \bl ] + \bphi_i) \in \BR^{d_h} \; \; , \nonumber
\end{align}
%
where ${[\cdot|\cdot]}$ is the concatenation of 2 vectors.
As shown in \cref{fig:design_exploration}, it fails to learn discernible shapes as it results in \bg{band-limiting the SDF with respect} to the latent space, which is neither desired nor relevant to our objectives.
%
%

\par \textbf{Design 2.}
Another approach is to append $\bl$ at every output layer, i.e. by updating \cref{eq:bacon} as ${f_{\Omega}(\bx,\bl,i) = \bW_i^{out} [\bz_i | \bl] + \bb_i^{out}}$.
As shown in \cref{fig:design_exploration}, cars are identifiable, but far from their ground truth and almost identical to one another.
We explain it by noticing that most of the network's computations are shared by all the different geometries, \bg{which have the same hidden layers $\bz_i$ values}.
As a result, this only adds a fixed SDF offset for each \bg{latent code} $\bl$. 
Equivalently, $\bl$ only varies the reconstructed level set of a scalar field that is fixed for all shapes.
%
%
\par \textbf{Design 3.}
The third approach is to append $\bl$ to all the hidden layers $\bz_i$.
As shown in \cref{fig:design_exploration}, this is the most effective solution, which we adopt.
More specifically, we concatenate $\bl$ to the sinus embeddings of spatial coordinates, and the network formulation in \cref{eq:bacon} becomes
%
\begin{align}
    \label{eq:latent_bacon}
    \bz_0 &= [\bbg_0(\bx) \;|\; \bl ] \mbox{ and } \bz_i = [\bbg_i(\bx) \;|\; \bl] \circ (\bW_i \bz_{i-1} + \bb_i) \nonumber \\ 
    f_{\Omega}(\bx,\bl,i) &= \bW_i^{out} \bz_i + \bb_i^{out} \;\; , \; 1 \leq i < N\; .
\end{align}
%
This architecture is graphically laid out in \cref{fig:pipeline}.
To accommodate for the addition of a latent vector, the dimension of the linear and output layers is increased from $d_h$ to $d_h + d_l$. 
For a fixed latent $\bl$, \cref{eq:latent_bacon} can be rewritten into \cref{eq:bacon} with simple matrix manipulations. 
As a result, each SDF $f_{\Omega}(\cdot,\bl,i)$ is bounded by $\sum_{j=0}^i B_j$ in the frequency domain, and this new latent architecture preserves the band-limiting property of \cref{eq:bacon}.
%
%
\subsection{Adjusting Bounded Frequencies} \label{sec:learn_frequency}
Throughout our experiments (see \cref{sec:ablations}), we found that the parameters of the coordinate embedding layers of \cref{eq:coord_embedding} play a crucial role for the overall surface detail, and even more importantly optimal parameters highly vary for different shape categories.
Thus, we optimize the frequencies $\bomega_i$ and phase shifts $\bphi_i$ instead of fixing them to their initial random uniform values of \cref{eq:sampling_freq}.
%
%
In order to maintain the band limiting properties of the network, coefficients of $\bomega_i$ must be restricted to the fixed range $[-B_i, B_i]$.
We thus, re-parametrize it as $\bomega_i = \tanh(\overline{\bomega_i}) * B_i$
and optimize $\overline{\bomega_i} \in \BR^{d_h \times 3}$ instead.
Coefficients $\bphi_i$ do not require such a re-parameterization and can be freely optimized, due to the periodicity of the $\sin(\cdot)$ function.
%
%
We add $\overline{\bomega_i}$ and $\bphi_i$ to the network's learnable parameters, which become 
${\Omega = \{\bW_i, \bb_i, \bW_i^{out}, \bb_i^{out}, \overline{\bomega_i}, \bphi_i \}}$. 
%
%
\subsection{Efficient Latent Space Exploration}\label{sec:exploration} \label{sec:method_meshing}
This network architecture provides important benefits for interactive latent space exploration.
Running the early layers of the network can provide a coarse, but smooth shape approximation.
To obtain the mesh $M_i(\bl)$ at level $i$ for a given latent code $\bl$, we apply Marching Cubes \cite{Lewiner03} to $f(\cdot,\bl,i)$. 
To accelerate iso-surface extraction, we use an octree structure, similar to \cite{Mescheder19,Venkatesh21,Guillard22a}. 
Initially, the SDF field $f(\cdot,\bl,i)$ is evaluated on a low-resolution regular grid (of shape $32^3$), which is then subdivided only for cells where the predicted value is smaller than the grid size. 
The SDF is subsequently queried at higher grid resolutions iteratively until we reach the desired resolution (typically $256^3$). 
This approach, combined with a low number of network layers, allows for rapidly obtaining a mesh, and \bg{real time} exploration of the latent space.
%
%
\par 
To increase the level of detail and manipulate the mesh $M_{i'}$ with $i' > i$, we use the same latent code $\bl$.
This differs from a naive solution employing multiple networks with various depths, which would not share their latent spaces. 
To accelerate iso-surface extraction for level $i'$, we re-employ SDF estimates from level $i$, following \cite{Lindell22}. 
We only query $f(\cdot,\bl,i')$ at grid voxels where $|f(\cdot,\bl,i)|$ is smaller than a threshold $\tau$.
This results in fewer operations than filling the octree from scratch with $f(\cdot,\bl,i')$, and, thus, a faster mesh recovery.
%
%
\subsection{Network Training and Inference} \label{sec:training}
We train our network in the auto-decoder fashion~\cite{Park20a}. 
Given a dataset $\mathcal{D}$ of watertight training shapes, we jointly optimize the network parameters $\Omega$ and a latent codebook $\bL \in \BR^{|\mathcal{D}| \times d_l}$ containing one code per shape. 
Training is performed by minimizing the sum of a loss on predicted SDF values $\mathcal{L}_\mathrm{sdf}$ and a regularization term $\mathcal{L}_\mathrm{reg}$ as $\min_{\Omega, \bL} \;\; \mathcal{L}_\mathrm{sdf} + \mathcal{L}_\mathrm{reg}.$
%
\par 
\textbf{SDF Loss.}
\bg{The SDF of the $j$-th training shape is sampled $S_f$ times close to the surface to provide \textit{fine} ground truth samples $(\dot{s}_{k}^{j}, \dot{\bx}_{k}^{j}) \in (\BR \times \BR^3)$ with $k=1,...,S_f$, and $S_c$ times further away from the surface for \textit{coarse} ground truth samples $(\ddot{s}_{k}^{j}, \ddot{\bx}_{k}^{j}) \in (\BR \times \BR^3)$. 
We sum the squared distances between ground truth and predicted SDFs}
%
\begin{align}
\label{eq:ldist}
\mathcal{L}_\mathrm{sdf} =  \left \| \dot{s}_{k}^{j} - f_{\Omega}(\dot{\bx}_{k}^{j}, \bL_j, i) \right \|_2^2  + \lambda_{c} \left \| \ddot{s}_{k}^{j} - f_{\Omega}(\ddot{\bx}_{k}^{j}, \bL_j, i) \right \|_2^2
\end{align}
%
at each network depth $i$, for each training shape $j$, and each SDF sample $k$.
We follow \cite{Lindell22,Duan20} and apply a lower weight $\lambda_{c} < 1$ to samples that are far from the surface.
Notice that all LoDs $i$ are supervised with the same ground truth SDF. 
No pre-processing (such as smoothing) is required for low LoDs, which are band-limited by the network architecture.
%
%
\par 
\textbf{Regularization Loss.}
As in \cite{Park20a}, we add an $L_2$ regularizer $\mathcal{L}_\mathrm{reg}$ on latent codes, with weight $\lambda_\mathrm{reg}$.
\par 
\textbf{Inference.} To test a network after training, we solve for the latent codes of unseen objects that best fit their SDFs.
The trained network parameters are frozen and act as a learned geometric prior \cite{Park20a}.
We optimize with gradient descent one latent code per unseen shape by minimizing $\mathcal{L}_\mathrm{sdf} + \mathcal{L}_\mathrm{reg}$ .

%
%
\section{Experiments}
We first demonstrate that our model is able to learn a parameterization of shapes that share geometric attributes and construct a coarse-to-fine hierarchy of details.
Secondly, we provide a quantitative comparison of our model to the most performant baselines, which demonstrates that our model is as accurate while exhibiting better behavior, \bg{and allowing to balance precision and computation time with a single network.}
Lastly, we perform an ablation study to understand the significance of optimizing the coordinate embedding parameters $\bomega_i$ and $\bphi_i$.


\begin{figure*}[t]
    \centering
    \begin{overpic}[width=0.97\textwidth]{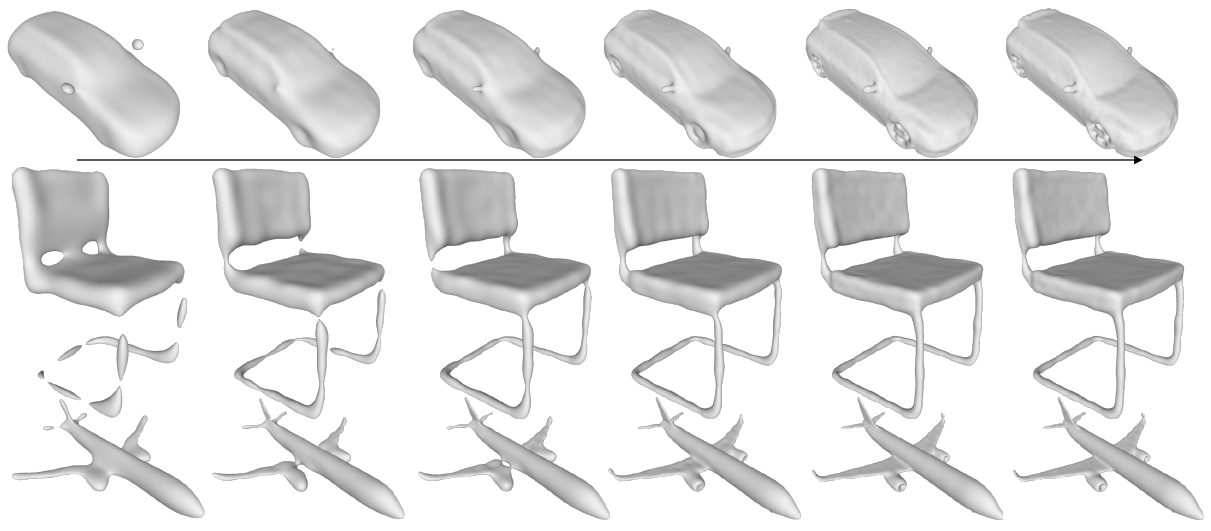}
        \put(3,31){$i=1$}
        \put(19,31){$i=2$}
        \put(36,31){$i=3$}
        \put(51,31){$i=5$}
        \put(66,31){$i=10$}
        \put(82,31){$i=12$}
    \end{overpic}
    %
    \vspace{-0.2cm}
    \caption{\textbf{Test Shape Reconstructions} 
    with varying levels of detail.
    %
    %
    The lowest possible level of detail ($i=1$) is smooth and already captures a very coarse structure of the shape.
    As $i$ increases, more details appear (up to saturation).
    }
    \label{fig:example}
    \vspace{-0.45cm}
\end{figure*}
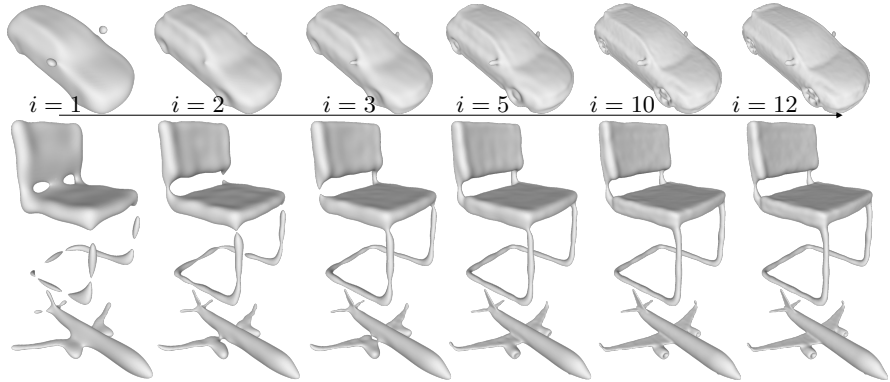

%
%
\subsection{Setting}
\par \textbf{Datasets.} 
We use 3 object categories from ShapeNet \cite{Chang15}: cars, chairs and airplanes, all normalized to fill 90\% of a $[-0.5, 0.5]^3$ bounding box.
For each of them, we train separate networks using 1200 shapes, and test them on 110 unseen shapes.

\par \textbf{Hyperparameters.}
We train networks with $N\!=\!13$ layers, resulting in $12$ SDF outputs of increasing frequencies.
We set the maximum spatial frequency to ${B = \sum_i B_i = 256}$, latent codes to a size of $d_l=256$ and coordinate embeddings to $d_h=256$.
For testing the network, we optimize an initial latent code $\bl=0 \in \BR^{d_l}$ for 1000 steps with Adam \cite{Kingma14a}, decreasing the learning rate linearly from $10^{-3}$ to $10^{-5}$ between steps 900 and 1000.
%
%
%
\subsection{Learning a Multiscale Shape Parameterization}
In \cref{fig:example}, we qualitatively demonstrate our model's ability to memorize training shapes and acquire a valuable prior to fit novel shapes.
The band-limited outputs progressively capture finer details throughout the network, leading to a learned hierarchy of details.
We observe that details seem to reach saturation for the last layers, which is further confirmed through quantitative assessment.
Despite not always preserving topology, low levels of detail capture the shape outlines.
\par
%

\begin{figure}[t]
\centering
\begin{minipage}{.49\textwidth}
    \centering
    \begin{overpic}[width=1.0\textwidth]{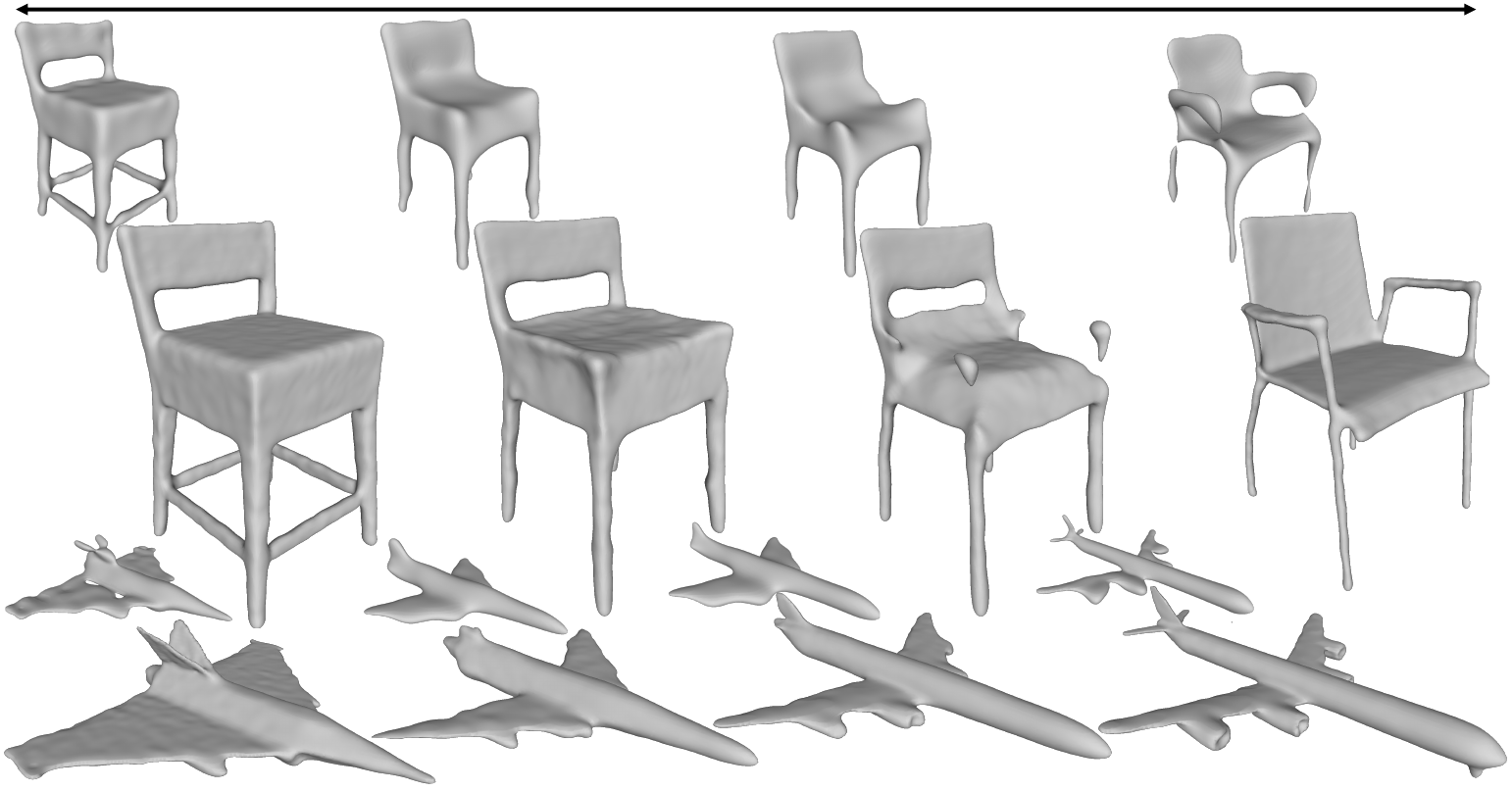}
            \put(11,48){$\bl_1$}
            \put(35,48){$\scriptstyle\tfrac{2}{3}\bl_1 + \tfrac{1}{3}\bl_2$}
            \put(60,48){$\scriptstyle\tfrac{1}{3}\bl_1 + \tfrac{2}{3}\bl_2$}
            \put(88,48){$\bl_2$}
    \end{overpic}
    \captionof{figure}{\textbf{Linear Interpolation}
        in the latent space translates into smooth shape deformations.
        Smaller objects correspond to a lower level of detail ($i=1$), and follow a similar transformation to the more detailed ones ($i=12$).}
    \label{fig:interp}
\end{minipage}%
\hfill
\begin{minipage}{.49\textwidth}
    \centering
    \includegraphics[width=1.0\textwidth]{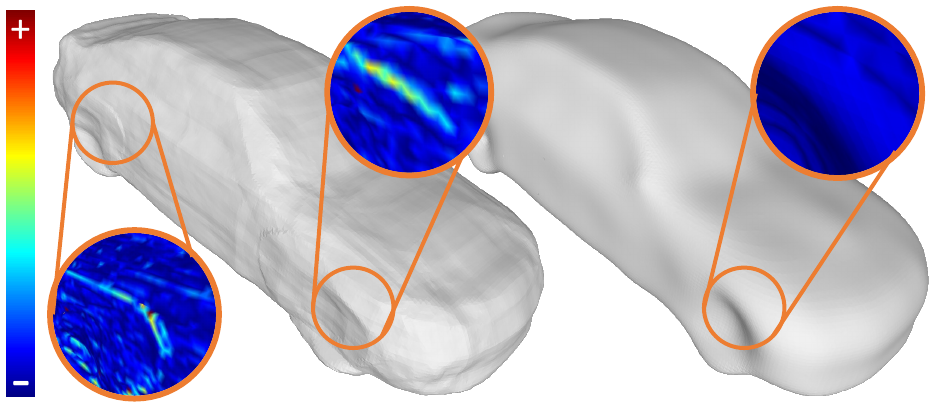}
    \captionof{figure}{
    \textbf{Surface Regularity (SR):} to measure regularity we average the vertex displacements that one step of Laplacian smoothing would cause.
        As shown in the insets, this captures surface cracks and bumps (left: 1 layer \baselineRelu{}, right: ours at $i=1$).}
    \label{fig:sr_metric}
\end{minipage}
\vspace{-0.5cm}
\end{figure}


\textbf{Interpolation in the Latent Space.}
\cref{fig:interp} illustrates the outcome of linear interpolation between two known latent codes
\bg{and shows} that the learned shape embedding is both continuous and complete.
Despite using unconventional activation functions on the latent codes, our novel architecture effectively represents meaningful shape abstractions,
and also allows for seamless topology changes.
%
%
\subsection{Shape Accuracy and Smoothness}
%


\textbf{Metrics.}
For measuring the faithfulness of the reconstructed surfaces, we use:
\begin{itemize}[noitemsep,topsep=0pt]
    \item The Chamfer Distance (CD) on 10k points on the reconstructed and ground truth surfaces, multiplied by $10^5$. Lower values mean higher  accuracy.
    \item The Earth mover's Distance (ED) for 5k surface points, using GeomLoss \cite{Feydy19} and multiplied by $10^4$. Again, lower distances imply better reconstruction.
\end{itemize}
For quantifying Surface Regularity (SR), we compute the average norm of each vertex displacement induced by one step of Laplacian smoothing, as shown in \cref{fig:sr_metric}.
The lower, the smoother the surface.
This metric is not computed with regards to a ground truth mesh, but is a per-shape score.
\par \textbf{Baselines.}
We compare our pipeline to other network architectures that rely on a compact latent code to represent shapes.
For them, we use another natural way of limiting the level of details, which consists in reducing the depth of a standard network, and, thus, its capacity.
We implement a ReLU network, with DeepSDF's architecture \cite{Park20a} and positional encoding~\cite{Tancik20} of degree 8, and refer to it as \baselineRelu{}.
We also implement a version of SIREN \cite{Sitzmann20}.
To learn geometric priors instead of overfitting a single shape, we condition it on latent codes via modulating layers, as in \cite{Dupont22}.
It uses sinus activations, and we refer to it as \baselineSin{}.
For each baseline, we train 12 networks, with depths ranging from 1 to 12.
We use the same hidden and latent sizes (512, 256) as for our network.
\par
%

\begin{table}[t]
  \small
  \centering
  \tabcolsep=0.03cm
  \begin{tabular}{cr|ccc|ccc|ccc|ccc}
   & Layer \#: & \multicolumn{3}{c}{2} & \multicolumn{3}{c}{4} & \multicolumn{3}{c}{8} & \multicolumn{3}{c}{12} \\
   & Metric& CD & ED & SR & CD & ED & SR & CD & ED & SR & CD & ED & SR \\
   \hline
   \parbox[t]{2mm}{\multirow{3}{*}{\rotatebox[origin=c]{90}{Cars}}} & {\baselineSin{}} &
   7.9  & 3.9 & 9.5 & 6.8  & 3.5 & 9.9 & 10.5 & 3.5 & 9.7 & 10.8 & 3.5 & 9.6 \\
   & {\baselineRelu{} } &
   \textbf{7.7} & \textbf{2.9} & 11.2 & \textbf{6.0} & \textbf{2.8} & 10.9 & \textbf{6.3} & \textbf{2.7} & 10.9 & \textbf{6.1} & \textbf{2.7} & 11.4 \\
   & Ours  &
   8.1 & 3.2 & \textbf{5.5} & 7.0 & 2.9 & \textbf{6.1} & \textbf{6.3} & 2.8 & \textbf{7.1} & 6.3 & 2.8 & \textbf{7.3} \\ \hline
   \parbox[t]{2mm}{\multirow{3}{*}{\rotatebox[origin=c]{90}{Chairs}}} & {\baselineSin{}} &
   14.9 & 5.0 & 13.0 & 13.4 & 4.6 & 12.0 & 10.7 & 4.9 & 16.0 & 12.7 & 6.4 & 19.0 \\
   & {\baselineRelu{}}  &
   \textbf{10.0} & \textbf{3.9} & 13.0 & \textbf{9.1} & \textbf{3.7} & 12.4 & 8.8 & 3.6 & 12.5 & 9.3 & 3.6 & 13.2  \\
   & Ours &
   14.3 & 5.4 & \textbf{7.8} & 10.1 & 4.0 & \textbf{7.8} & \textbf{8.3} & \textbf{3.4} & \textbf{8.3} & \textbf{8.3} & \textbf{3.4} & \textbf{8.4}  \\ \hline
   \parbox[t]{2mm}{\multirow{3}{*}{\rotatebox[origin=c]{90}{Planes}}} & {\baselineSin{} } &
   4.5 & 3.3 & 14.0 & 3.6 & 2.5 & 13.7 & 2.9 & 2.1 & 14.7 & 5.2 & 3.4 & 16.1 \\
   & {\baselineRelu{} }  &
   \textbf{3.0} & \textbf{2.0} & 15.0 & \textbf{2.7} & \textbf{2.0} & 14.5 & \textbf{2.6} & 1.8 & 15.5 & \textbf{2.4} & 1.6 & 14.8  \\
   & Ours  &
   5.8 & 3.0 & \textbf{12.2} & 3.8 & 2.3 & \textbf{13.0} & 2.7 & \textbf{1.5} & \textbf{14.1} & 2.7 & \textbf{1.5} & \textbf{14.5}  \\
  \end{tabular}
  \vspace{1mm}
  \caption{\textbf{Unseen Test Shapes}: Chamfer Distances (CD), Earth mover's Distances (ED) and Surface Regularity (SR) -- lower values indicate better results. We achieve comparable best (CD, ED) while being smoother (SR) than the best baseline (\baselineRelu{}). \baselineSin{} and \baselineRelu{} require training one network per category and LoD ours simply needs per-category training.}
  \label{tab:testset_accuracy}
  \vspace{-8mm}
  
\end{table}


%

\begin{figure}
\centering
\begin{minipage}{.6\textwidth}
    \centering
    \includegraphics[width=1.0\textwidth]{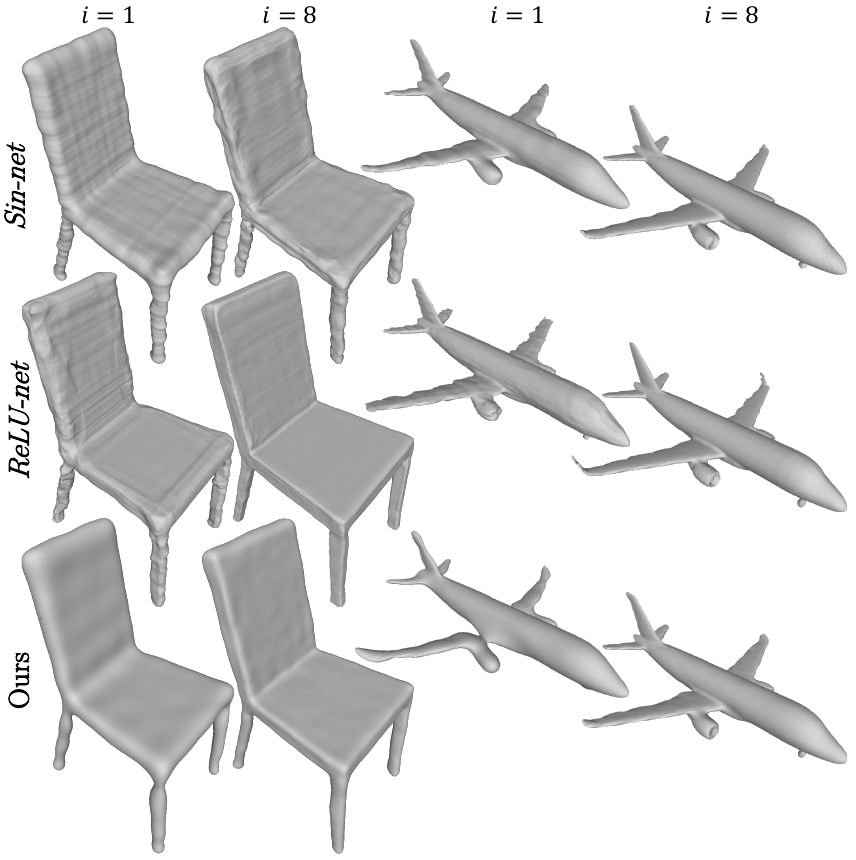}
    \vspace{-0.7cm}
    \captionof{figure}{\textbf{Unseen Test Shapes}
    as reconstructed by our model, \baselineSin{} and \baselineRelu{}.
    It is visible on flat regions of the chair that our model provides smoother surfaces at lower LoD ($i=1$).}
    \label{fig:comparison}
\end{minipage}%
\hfill
\begin{minipage}{.38\textwidth}
    \centering
    \includegraphics[width=.9\textwidth]{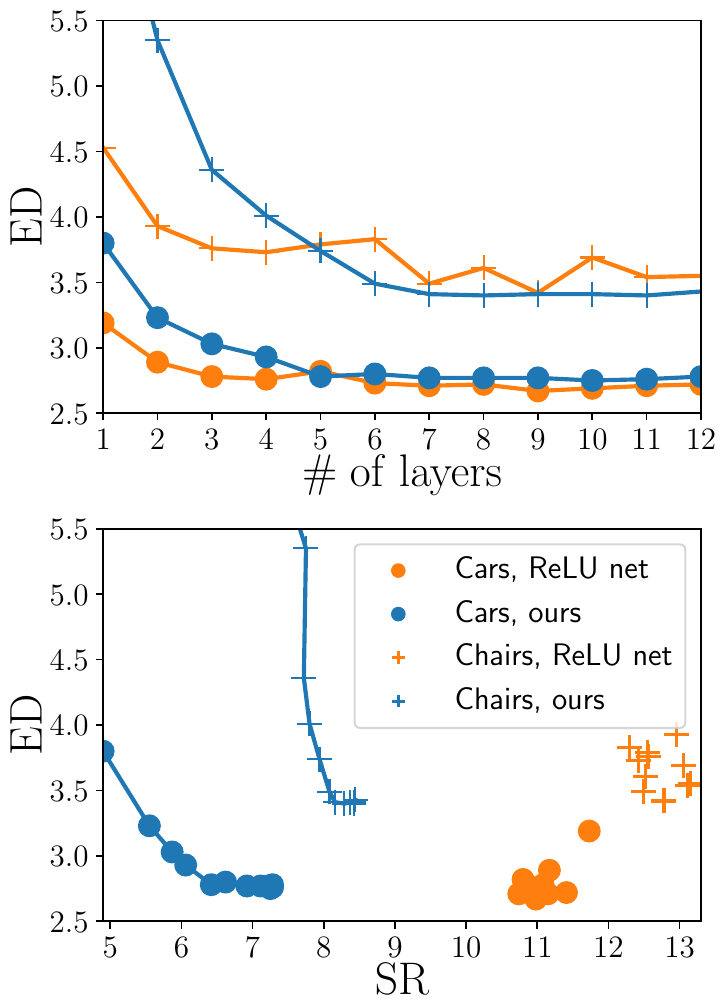}
    \vspace{-0.5cm}
    \captionof{figure}{\textbf{Earth mover's Distance (ED)}. 
    \textbf{Top:} ED as a function of depth.
    \textcolor{plot_orange}{\textit{ReLU-nets}} (1 per depth) are less monotonous than \textcolor{plot_blue}{ours} (single network).
    \textbf{Bottom:} 
    \textcolor{plot_blue}{Ours} exhibits a clear tradeoff between ED/SR, but not \textcolor{plot_orange}{\textit{ReLU-nets}}.}
    \label{fig:plot}
\end{minipage}
    \vspace{-0.5cm}
\end{figure}

\textbf{Accuracy.}
\cref{tab:testset_accuracy} compares our network with two baseline models, \baselineRelu{} and \baselineSin{}, for depths $i=2,4,8,12$ on the test splits for all three shape categories.
Our method achieves minimal ED and CD scores that are very similar to \baselineRelu{} and consistently better than \baselineSin{}.
This validates that our network representation of 3D surfaces is comparable to the best baselines in terms of accuracy.
\cref{fig:comparison} shows a qualitative comparison.
\par
In \cref{fig:plot} (top), we plot reconstruction accuracy (ED) as a function of network depth for our single network and 12 \baselineRelu{} networks of varying depths for the cars and chairs datasets.
The error monotonously decreases with depth for our solution, which is not the case for \baselineRelu{}.
Therefore, band-limiting makes our solution more consistent and predictable. 
This also highlights another practical benefit of a single network approach: it can be pruned after training.
By discarding layers $i\! >\! 8$, we avoid manual hyperparameter tuning and get the best performance with a single training.
\par
It is worth noting that \baselineRelu{} and \baselineSin{} require one network for each depth, resulting in one latent code per depth for encoding each shape.
It could imaginably be possible to train these models with a shared latent space, or with one SDF head per layer.
They would however still exhibit a greater degree of surface irregularity, which we explain in more detail below.
\par \textbf{Surface Regularity.}
Across all object categories and depths, our network consistently achieves lower SR values, as shown in \cref{tab:testset_accuracy}.
This indicates that the shapes generated by our network are always smoother than those generated by \baselineRelu{} and \baselineSin{}. 
This trend holds at higher levels of detail, with almost equal ED and CD values.
\par
Moreover, our pipeline allows for an explicit and controllable trade-off between shape accuracy (ED) and surface regularity (SR).
As demonstrated in \cref{fig:plot} (bottom), there is a very consistent trend: as the output layer becomes deeper, accuracy improves while smoothness decreases.
In contrast, \baselineRelu{} does not exhibit such a correlation.
%
%
\subsection{Shape Reconstruction Speed}

\begin{figure}[t]
\tabcolsep=0.1cm 
\begin{center}
\begin{tabular}{ccc}
\includegraphics[width=0.26\textwidth, trim={1cm 0 1cm 0},clip]{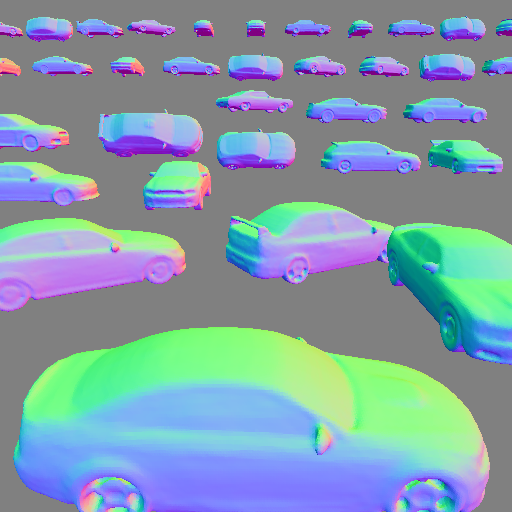} &
\includegraphics[width=0.26\textwidth, trim={1cm 0 1cm 0},clip]{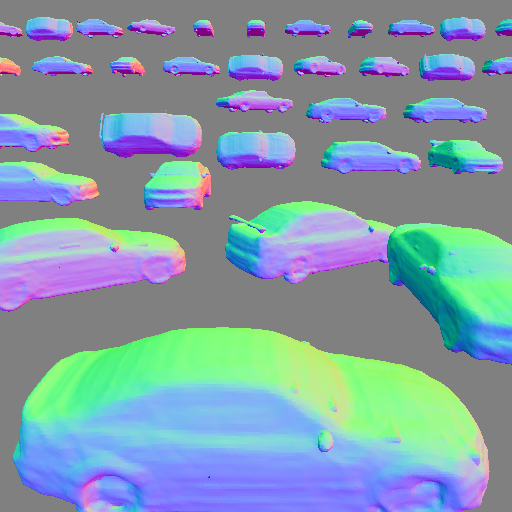} &
\includegraphics[width=0.26\textwidth, trim={1cm 0 1cm 0},clip]{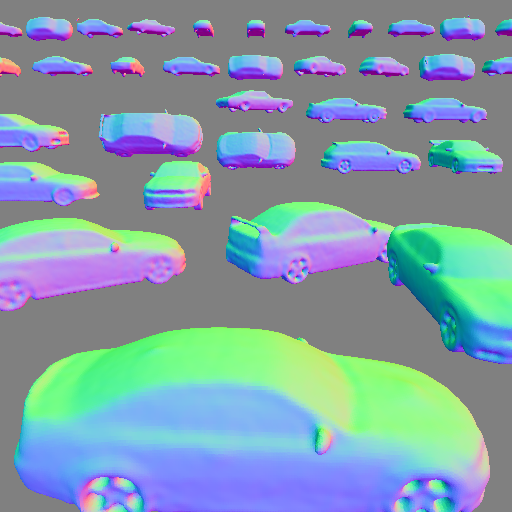} \\
{\small \textbf{(a)} 4.9 sec., with}&
{\small \textbf{(b)} 1.0 sec., with }&
{\small \textbf{(c)} 1.7 sec., ours,}\\
{\small 10-layers \textit{ReLU-net}.}&
{\small 2-layers \textit{ReLU-net}.}&
{\small variable depth.}\\
\end{tabular}
\end{center}
\vspace{-0.5cm}
\caption{\bg{
\textbf{Meshing and rasterizing implicit shapes:}}
\bg{\textbf{(a)} A deep \textit{ReLU-net} provides details but is slow.
\textbf{(b)} A shallower \textit{ReLU-net} is fast but coarse.
\textbf{(c)} With ours, network depth and LoD can be adjusted based on camera distance, to render quickly and without artifacts.}
}
\label{fig:scene_rendering}
\vspace{-0.45cm}
\end{figure}

At level $i=8$, reconstructing a mesh at resolution 256 using the octree subdivision method described in \cite{Mescheder19,Venkatesh21,Guillard22a} takes an average of \textbf{66.7ms} on an Nvidia A100 GPU, \bg{of which only 3.4ms is spent on Marching Cubes \cite{Yatagawa20}.}
It is reduced to \textbf{58.9ms} by early stopping the network computations when $|f(\cdot, \cdot,i')| > \tau$ with $i'<8$, as suggested in \cite{Lindell22} and explained in \cref{sec:method_meshing}.
This reduction in computation time is a significant advantage of the multiscale network architecture, which provides coarse SDF estimates in just a few operations.
\par
\bg{Furthermore, our network can adaptively balance computation time and precision. 
This proves useful for meshing and rasterizing a collection of implicit objects, and rendering an image of a 3D scene as depicted in \cref{fig:scene_rendering}.
By using coarse LoD ($i<4$) for distant objects while maintaining high quality for nearby ones, we can render the scene much faster than a \textit{ReLU-net}.}
%
%
\subsection{Ablations}
\label{sec:ablations}

\begin{wrapfigure}{l}{0.25\textwidth}
 \vspace{-0.8cm}
  \begin{tabular}{r|cc}
            & CD             & ED \\ \hline
  Ours      &  \textbf{8.3}  &  \textbf{3.4}  \\
  \arrayrulecolor{gray}\hline\arrayrulecolor{black}
  Fixed $\bomega$ 
            &  8.9          &  3.6  \\
  Fixed $\bphi$ 
            &  8.7          &  3.5  \\
   \begin{tabular}[r]{@{}r@{}}Fixed $\bomega$ \vspace{-1mm} \\ and $\bphi$\end{tabular}
            &  9.0          &  3.5
  \end{tabular}

\label{tab:ablation}
\vspace{-1cm}
\end{wrapfigure}



%
We perform an ablation on the optimization of the coordinate embedding parameters $\bomega$ and $\bphi$ in our approach.
We analyze the performance on test chairs at a depth of $i=8$ when these parameters are left fixed to their initial values.
As shown in the inset table, there is a moderate drop in performance when these parameters are not optimized.

In another experiment, we trained on chairs, but freezing $\bomega$ and $\bphi$ to the ones obtained by training on cars. This resulted in a 15\% increase of ED, again demonstrating the importance of learning category specific frequencies and phases.
%
%
\subsection{Shape Completion}

\begin{wrapfigure}{r}{0.5\textwidth}
    \vspace{-1cm}
    \centering
    \begin{overpic}[width=0.49\textwidth]{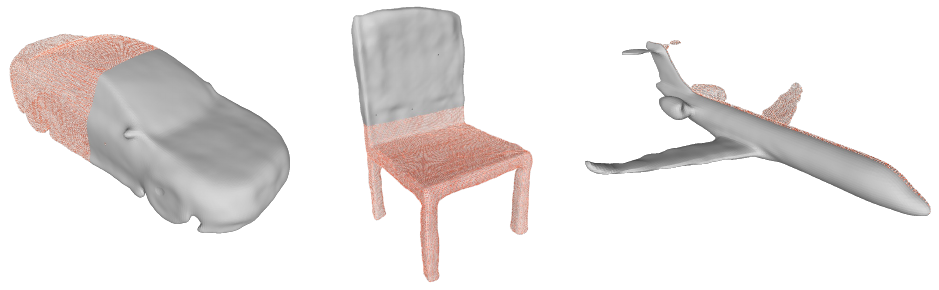}
    \end{overpic}
    \vspace{-0.5cm}
    \caption{\textbf{Shape Completion}
    Our network can complete unseen 3D shapes even when the SDF signal is incomplete. In these examples, only the red half of the object received SDF supervision, yet the compact latent code parameterization yields entire shapes.
    }
    \label{fig:completion}
    \vspace{-0.6cm}
\end{wrapfigure}

In \cref{fig:completion}, we showcase our network's capability to reconstruct shapes based on partial SDF signals.
The gradient descent procedure we employ is similar to the one used for the previous section, but this time the SDF supervision is limited to only half of the 3D domain.
Despite the reduced supervision, our pipeline successfully recovers valid shapes that adhere to the provided SDF signal, while also generating plausible completions in regions lacking supervision.
By employing a single latent code to parameterize the entire 3D distance field, our network acts a strong shape prior.
%


\section{Conclusion}
We propose a novel latent conditioned architecture for implicit 3D shape representation with multiple levels of details, and enable explicit control of the SDF maximum frequency for each level.
This architecture is a learned shape parameterization from a single latent vector.
Our experiments demonstrate that, at the highest level of detail, it is just as accurate as other existing methods, while consistently producing smoother results with fewer surface artifacts at lower levels of detail.
Additionally, it offers a controllable tradeoff between \bg{speed} and accuracy.
\par
We will investigate a disentangled latent representation for each LoD, which would enable mix-and-matching coarse outlines and local details.
This would allow for greater flexibility and control over the generation of 3D shapes.

\section*{Acknowledgments}
This work was supported in part by the Swiss National Science
Foundation.
Christian Theobalt was supported by ERC Consolidator Grant 4DReply (770784).

%
%

%
%
%
%
\bibliographystyle{splncs04}
\bibliography{string,egbib}

\clearpage


\section{Proof of the band-limits of our network}

Our network with $N$ layers maps a latent code $\bl \in \BR^{d_l}$ and an input coordinate $\bx \in \BR^3$ to $N - 1$ estimates of the SDF at this position, $f_{\Omega}(\bx, \bl, i)$ with $1 \leq i < N$. 
The input coordinate $\bx \in \BR^3$ is first encoded using sine layers 
%
\begin{align}
\label{eq:supp_coord_embedding}
    \bbg_i(\bx)=\sin(\bomega_i \bx + \bphi_i) \in \BR^{d_h}
\end{align}
%
for $i=0,...,N-1$, with $\bomega_i \in \BR^{d_h \times 3}$, $\bphi_i \in \BR^{d_h}$,  and $d_h$ the hidden dimension.
Intermediate activations $\bz_i \in \BR^{d_h+d_l}$ and SDF outputs $f_{\Omega}(\bx,\bl,i) \in \BR$ are defined as
%
\begin{align}
    \label{eq:supp_latent_bacon}
    \bz_0 &= [\bbg_0(\bx) \;|\; \bl ]\nonumber \\ 
    \bz_i &= [\bbg_i(\bx) \;|\; \bl] \circ (\bW_i \bz_{i-1} + \bb_i) \nonumber \\ 
    f_{\Omega}(\bx,\bl,i) &= \bW_i^{out} \bz_i + \bb_i^{out} \;\; , \; 1 \leq i < N\; .
\end{align}
%
where $\circ$ is the Hadamard product, ${(\bW_i, \bb_i) \in \BR^{(d_h+d_l) \times (d_h+d_l)} \times \BR^{d_h+d_l}}$ are the \textit{linear layers} and ${(\bW_i^{out}, \bb_i^{out}) \in \BR^{1 \times (d_h+d_l)} \times \BR}$ are the \textit{output layers}.

We here show that for a fixed latent code $\bl \in \BR^{d_l}$, the above formulation can be rewritten into a standard \textsc{Bacon} network~\cite{Lindell22}, and thus enjoys the same band-limits.
The main intuition is to hardcode $\bl$ as a fixed component within the coordinate embeddings.

Let us first define $\alpha = \max(|\bl|)$ the largest coefficient of $\bl$ in absolute scale, thus allowing us to write 
$$
\bl = \alpha \sin( \overline{\bl} )
$$
for some $\overline{\bl} \in \BR^{d_l}$. Using this expression of $\bl$ we define $\bbg_i'(\cdot)$ as
$$
\bbg_i'(\bx) = \sin( [\bomega_i \;|\; 0] \bx + [\bphi_i \;|\; \overline{\bl}])
$$
where $[\bomega_i \;|\; 0] \in \BR^{(d_h + d_l) \times 3}$ means the concatenation of $\bomega_i$ with $d_l$ zeros vectors $(0,0,0)$ along the row axis. By denoting $[1 \;|\; \alpha] \in \BR^{d_h + d_l}$ the concatenation of $d_h$ coefficients $1$ followed by $d_l$ coefficients $\alpha$, we get
$$
[\bbg_i(\bx) \;|\; \bl] = [1 \;|\; \alpha] \circ \bbg_i'(\bx) \; \; \;.
$$

The multiplication by $[1 \;|\; \alpha]$ can by absorbed in the matrix multiplications. We define $\bW_i'$ and $\bW_i'^{out}$ by scaling the last $d_l$ columns of $\bW_i$ and $\bW_i^{out}$ by $\alpha$:
\begin{align}
    \bW_i' &= \bW_i \circ \begin{pmatrix}
                            [1 \; | \; \alpha]\\ 
                            ...\\ 
                            [1 \; | \; \alpha]
                            \end{pmatrix} \\ 
    \bW_i'^{out} &= \bW_i^{out} \circ \begin{pmatrix}
                            [1 \; | \; \alpha]\\ 
                            ...\\ 
                            [1 \; | \; \alpha]
                            \end{pmatrix}
\end{align}
As a result, by fixing $\bl$, \Cref{eq:supp_latent_bacon} can be rewritten as a standard \textsc{Bacon} network~\cite{Lindell22} with hidden dimension $d_h + d_l$ and parameters
\begin{align}
    \bomega_i' &= [\bomega_i \;|\; 0] \;, \\ 
    \bphi_i' &= [\bphi_i \;|\; \overline{\bl}] \;, \\ 
    \bW_i' &= \bW_i \circ \begin{pmatrix}
                            [1 \; | \; \alpha]\\ 
                            ...\\ 
                            [1 \; | \; \alpha]
                            \end{pmatrix} \;, \\ 
    \bb_i' &= \bb_i \;, \\ 
    \bW_i'^{out} &= \bW_i^{out} \circ \begin{pmatrix}
                            [1 \; | \; \alpha]\\ 
                            ...\\ 
                            [1 \; | \; \alpha]
                            \end{pmatrix} \;, \\ 
    \bb_i'^{out} &= \bb_i^{out}  \;.
\end{align}


\section{Hyperparameters}
We set the maximum spatial frequency to ${B = \sum_i B_i = 256}$ and distribute the bandwidth as
\begin{itemize}
  \item ${B_0 = B / 48}$ ; 
  \item ${B_1 = B_2 = B_3 = B_4 = B / 24}$ ;
  \item ${B_5 = B_6 = B_7 = B / 16}$ ;
  \item ${B_8 = B_9 = B_{10} = B_{11} = B_{12} = B / 8}$.
\end{itemize}
Latent codes are of size $d_l=256$ and coordinate embeddings of size $d_h=256$, for a total layer width of $512$.
Our networks are trained for $10^6$ steps with the Adam optimizer \cite{Kingma14a}, and with a learning rate decreasing logarithmically from $10^{-2}$ to $10^{-4}$.
We use mini-batches of 4 shapes, each of them represented as 10k SDF samples at each iteration.
We set the coarse sample weights to $\lambda_c=10^{-2}$ in, and the regularizer weight to $\lambda_{reg}=10^{-4}$.

We follow the SDF sampling strategy of \cite{Park20a} for querying ground truth samples.
The fine samples used in $\mathcal{L}_{sdf}$ are the top 5\% that are the closest to the surface, remaining ones are the coarse ones.

\bg{The renderings of Fig. 8 were done using the mesh rasterization modules of PyTorch3D~\cite{ravi2020pytorch3d}, with disabled backpropagation and at a resolution of 512*512 pixels.}


\section{Adding Eikonal regularization to \baselineRelu{}}

\begin{figure}
    \centering
    \includegraphics[width=0.8\textwidth]{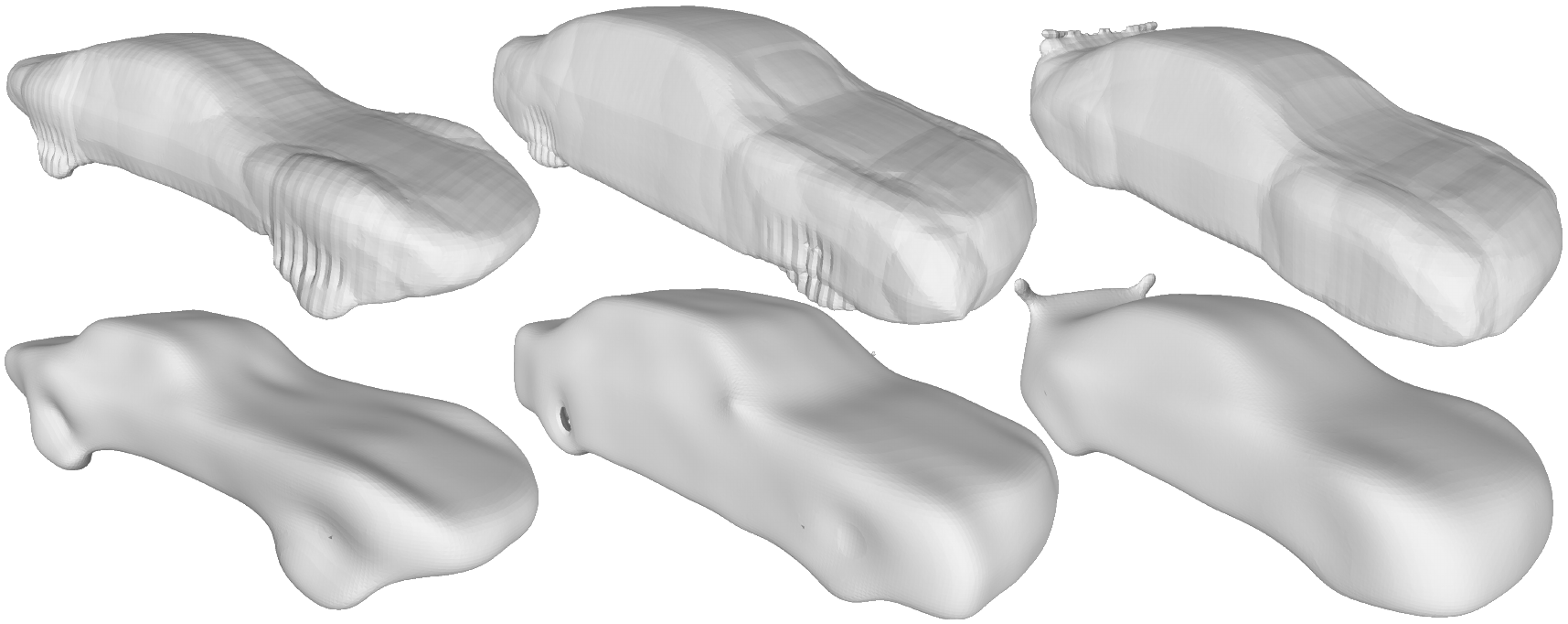}
    %
    \caption{\textbf{Eikonal regularization}
    on a 1-layer \baselineRelu{} (\textbf{top}) still yields rough surfaces with many artefacts compared to our network at level $i=1$ (\textbf{bottom}).
    }
    \label{fig:eikonal}
\end{figure}

We added an Eikonal regularization \cite{Atzmon20b} to a shallow \baselineRelu{} network to test its effect on surface regularity.
It consists in (softly) constraining the spatial gradient norms to be equal to 1 --- as for a true SDF field.
As shown in \Cref{fig:eikonal}, this does not fix surface artefacts.
In addition, it is significantly slower to train because it requires the computation of second order gradients.


\section{Setting the maximum bandwidth $B$}

\begin{figure*}
    \centering
    \includegraphics[width=0.8\textwidth]{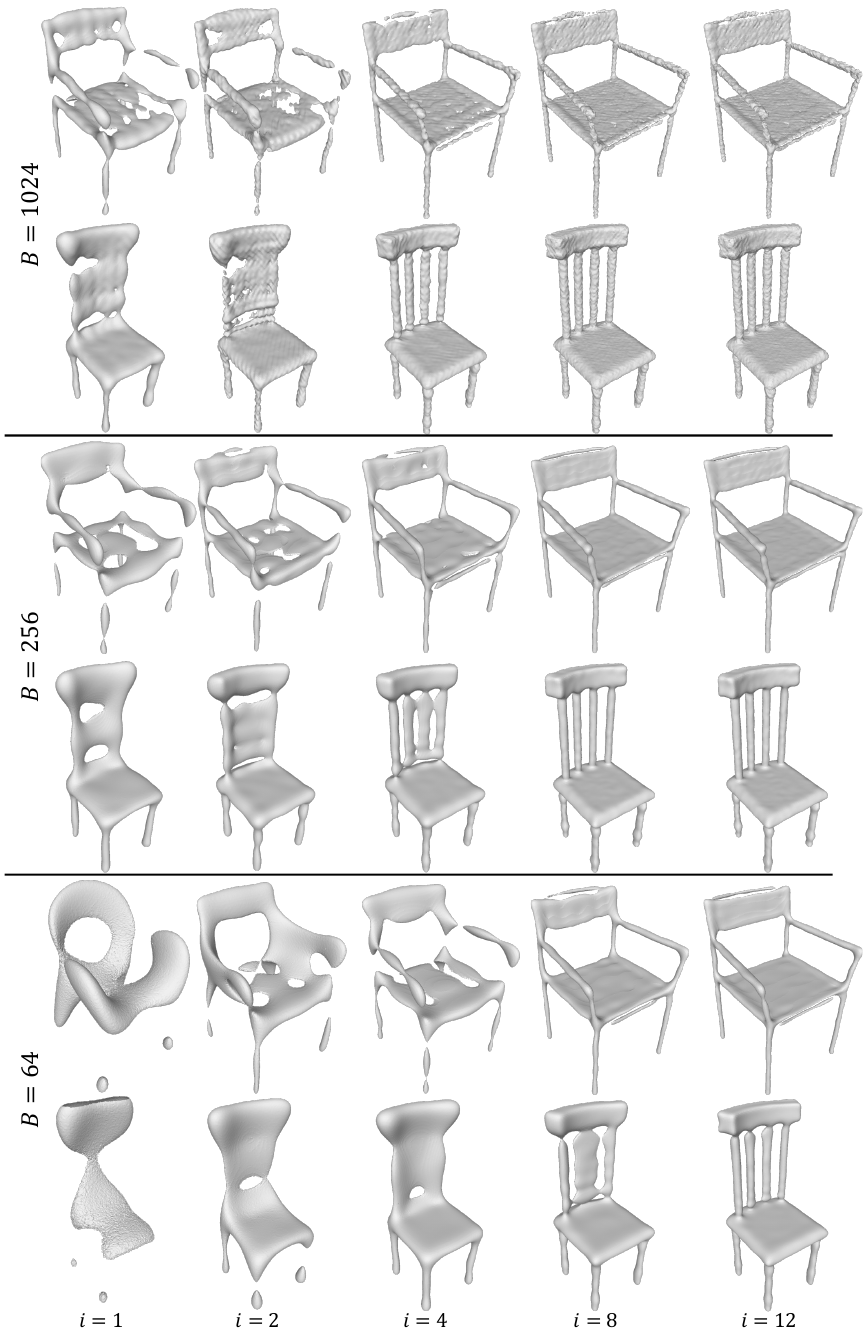}
    %
    \caption{\textbf{Setting the maximum bandwidth $B$:} 
    we compare reconstructed chairs with a high maximum frequency ($B=1024$, \textbf{top}), our default value ($B=256$, \textbf{middle}), and a low maximum frequency ($B=64$, \textbf{bottom}), at different levels of detail $i$. With $B=1024$, surfaces are not smooth and show high frequency artefacts. With $B=64$, the details increase more gradually, but the highest level of detail still shows over-smoothed shapes.
    }
    \label{fig:change_B}
\end{figure*}


\begin{table}[]
  \small
  \centering
  \tabcolsep=0.03cm
  \begin{tabular}{r|ccc|ccc|ccc|ccc}
   Layer \#: & \multicolumn{3}{c}{2} & \multicolumn{3}{c}{4} & \multicolumn{3}{c}{8} & \multicolumn{3}{c}{12} \\
   Metric& CD & ED & SR & CD & ED & SR & CD & ED & SR & CD & ED & SR \\
   \hline
   \multicolumn{1}{l|}{\cellcolor[HTML]{C0C0C0}Ours with }    & \multicolumn{12}{|l}{\cellcolor[HTML]{C0C0C0}Chairs}  \\
   $B=64$ &
   30.8 & 8.0 & \textbf{5.6} & 22.3 & 6.7 & \textbf{4.9} & 11.3 & 4.1 & \textbf{7.5} & 11.3 & 4.0 & \textbf{7.9} \\
   $\mathbf{B=256}$  &
   \textbf{14.3} & 5.4 & 7.8 & 10.1 & 4.0 & 7.8 & 8.3 & 3.4 & 8.3 & 8.3 & 3.4 & 8.4  \\
   $B=1024$ &
   14.9 & \textbf{5.2} & 12.6 & \textbf{8.7} & \textbf{3.6} & 11.7 & \textbf{8.2} & \textbf{3.4} & 15.9 & \textbf{8.2} & \textbf{3.4} & 17.9 \\
  \end{tabular}
  \vspace{1.5mm}
  \caption{\textbf{Unseen test chairs}: Chamfer Distances (CD), Earth mover's Distances (ED) and Surface Roughness (SR) for varying maximum bandwidth values $B$ (by default $B=256$ in our paper).
  With $B=64$ surfaces are smoother (lower SR) but less precise (higher CD and ED).
  With $B=1024$, they fit more closely the ground truth (lower CD and ED) but show many local artefacts (higher SR), as shown on \Cref{fig:change_B}.}
  \label{tab:change_B}
\end{table}

In \Cref{fig:change_B,tab:change_B} we train variants our network on chairs with a maximum bandwidth $B=64$, $B=1024$, and compare to our default value of $B=256$ on the test set.
With a lower bandwidth, the level of detail increases more gradually, but reconstruction accuracy is decreased.
A higher bandwidth leads to metrics saturating quicker, and thus a less progressive increase in details, as well as high frequency artefacts on the reconstructed surfaces.


\begin{figure*}
    \centering
    \includegraphics[width=0.95\textwidth]{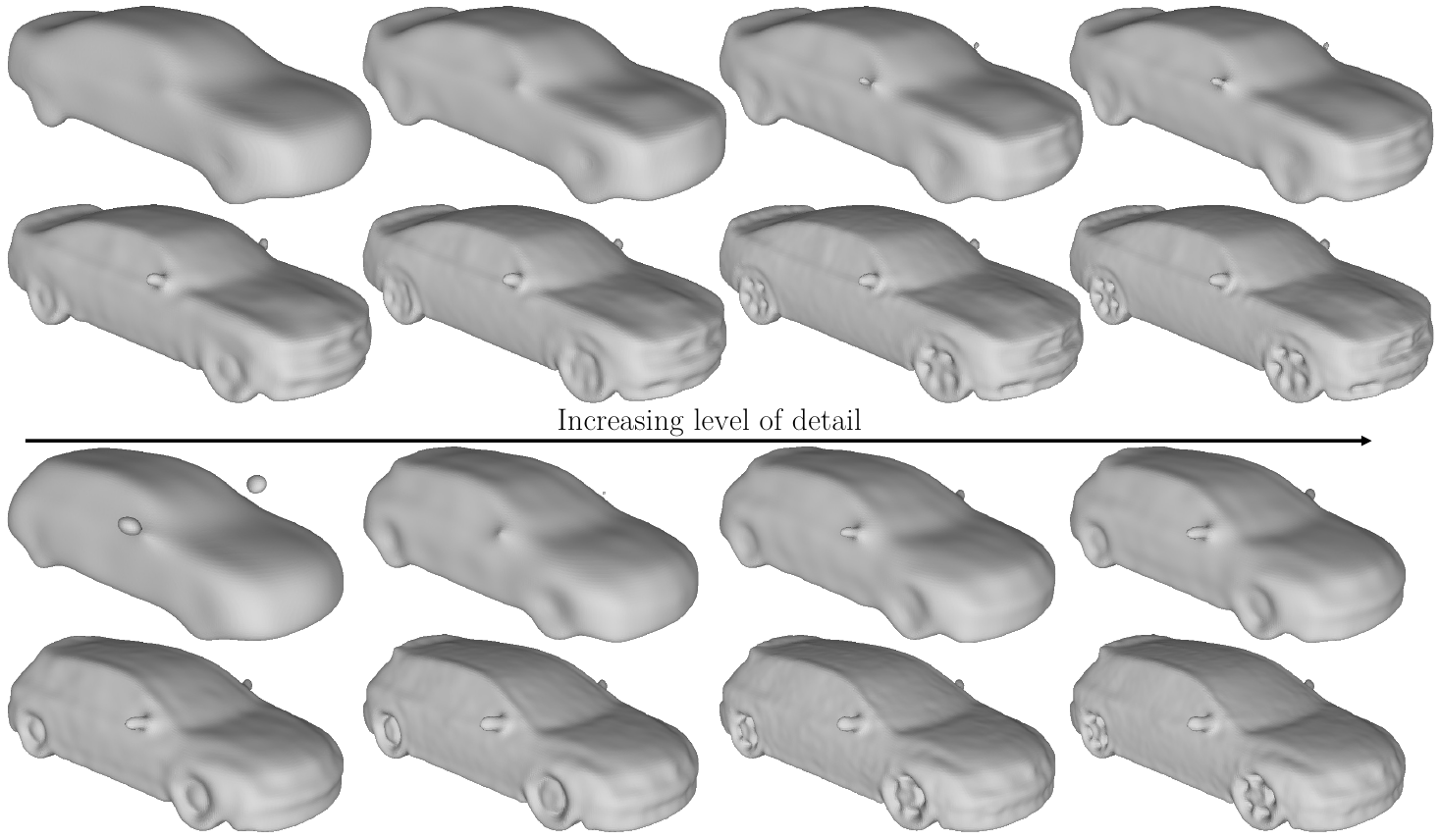}
    %
    \caption{\textbf{Increasing level of detail} with our model on cars, from $i=1$ to $i=8$.
    }
    \label{fig:increase_lod}
\end{figure*}

In \Cref{fig:increase_lod} we show and example of increasing the level of detail with our network when $B=256$, with ${B_0 = B / 48}$ ; ${B_1 = B_2 = B_3 = B_4 = B / 24}$ ; ${B_5 = B_6 = B_7 = B / 16}$ and ${B_8 = B_9 = B_{10} = B_{11} = B_{12} = B / 8}$.


\section{Multi-scale \baselineRelu{}}

\begin{table}[]
  \small
  \centering
  \tabcolsep=0.03cm
  \begin{tabular}{r|ccc|ccc|ccc|ccc}
   Layer \#: & \multicolumn{3}{c}{2} & \multicolumn{3}{c}{4} & \multicolumn{3}{c}{8} & \multicolumn{3}{c}{12} \\
   Metric& CD & ED & SR & CD & ED & SR & CD & ED & SR & CD & ED & SR \\
   \hline
   
   \multicolumn{1}{l|}{\cellcolor[HTML]{C0C0C0} }    & \multicolumn{12}{|l}{\cellcolor[HTML]{C0C0C0}Chairs}  \\
   {\begin{tabular}[c]{@{}r@{}}\textit{MS-ReLU} \vspace{-1.7mm} \\ \textit{\scriptsize{(single net.)}} \vspace{-1.2mm} \end{tabular}} &
   17.4 & 5.5 & 10.1 & \textbf{8.7} & \textbf{3.6} & 10.2 & 8.6 & \textbf{3.4} & 10.3 & 8.5 & \textbf{3.4} & 11.4 \\
   {\begin{tabular}[c]{@{}r@{}}\textit{ReLU-net} \vspace{-1.7mm} \\ \textit{\scriptsize{(1 / column)}} \vspace{-1.2mm} \end{tabular}}  &
   \textbf{10.0} & \textbf{3.9} & 13.0 & 9.1 & 3.7 & 12.4 & 8.8 & 3.6 & 12.5 & 9.3 & 3.6 & 13.2  \\
   \begin{tabular}[c]{@{}r@{}}Ours \vspace{-1.7mm} \\ \textit{\scriptsize{(single net.)}} \vspace{-0.6mm} \end{tabular} &
   14.3 & 5.4 & \textbf{7.8} & 10.1 & 4.0 & \textbf{7.8} & \textbf{8.3} & \textbf{3.4} & \textbf{8.3} & \textbf{8.3} & \textbf{3.4} & \textbf{8.4}  \\
   
  \end{tabular}
  \vspace{1.5mm}
  \caption{\textbf{Unseen test chairs}: Chamfer Distances (CD), Earth mover's Distances (ED) and Surface Roughness (SR). Lower values indicate better results for all metrics.
  \textit{MS-ReLU}, which is a multiscale version of of \baselineRelu{}, performs slightly better than the latter.
  Our network is however consistently smoother (SR) while achieving comparable accuracy (CD, ED).
  }
  \label{tab:supp_testchairs_ms_relu}
\end{table}

We extended \baselineRelu{} into a multi-scale version. In the spirit of our own architecture, we add one output layer per hidden layer.
The resulting network thus provides multiple SDFs outputs from a single network and latent code, each corresponding to a depth level.
We refer to this method as \textit{MS-ReLU}.
It solves one issue of \baselineRelu{} and \baselineSin{}, by mapping a single latent code to  multiple surface estimates, without requiring to put different latent spaces in correspondence.

However, it still suffers from the same artefacts as \baselineRelu{} with rougher surfaces, as reported in \Cref{tab:supp_testchairs_ms_relu} on test chairs.
Moreover, it provides no explicit control of the bandwidth, and the coarse to fine hierarchy is only tied to the increasing network capacity.


\section{Decreasing Marching Cubes Resolution}
Decreasing marching cubes resolution is a trivial baseline for reducing the level of detail and computation time. However, as shown in \Cref{tab:mc_res_decrease}, this solution negatively impacts the reconstruction accuracy. A grid coarse grid resolution yields very inaccurate shapes. By comparison, our method provides better shape approximations.

\begin{table*}[]
  \centering
  
\begin{tabular}{llcc}
      & \multicolumn{1}{l|}{}  & \multicolumn{1}{c}{MC=$64^3$} & \multicolumn{1}{c}{MC=$256^3$} \\ \cline{2-4}
Layer \# :  & \multicolumn{1}{l|}{1} &            2.94            &          2.89               \\
            & \multicolumn{1}{l|}{2} &            1.54            &          1.26                \\
            & \multicolumn{1}{l|}{3} &            1.46            &          1.02                \\
            & \multicolumn{1}{l|}{4} &            1.47            &          0.90                \\
            & \multicolumn{1}{l|}{5} &            1.46            &          0.84                \\
            & \multicolumn{1}{l|}{6} &            1.68           &           0.79                \\
            & \multicolumn{1}{l|}{7} &            1.78            &          0.77                \\
            & \multicolumn{1}{l|}{8} &            1.89            &          0.75             \\
            & \multicolumn{1}{l|}{9} &            1.93            &          0.74                \\
            & \multicolumn{1}{l|}{10} &           1.90             &         0.72               \\
            & \multicolumn{1}{l|}{11} &           1.91             &         0.72               \\
            & \multicolumn{1}{l|}{12} &           1.94             &         0.72               
\end{tabular}

\caption{
\textbf{Chamfer Distance (CD) with marching cubes resolutions (MC) of 64 and 256}, on the chairs dataset. A low marching cubes resolution always yield more inaccurate reconstructions.
}
\label{tab:mc_res_decrease}
\end{table*}


\section{Additional Qualitative Results}

\begin{figure*}[t]
    \centering
    \begin{overpic}[width=0.97\textwidth]{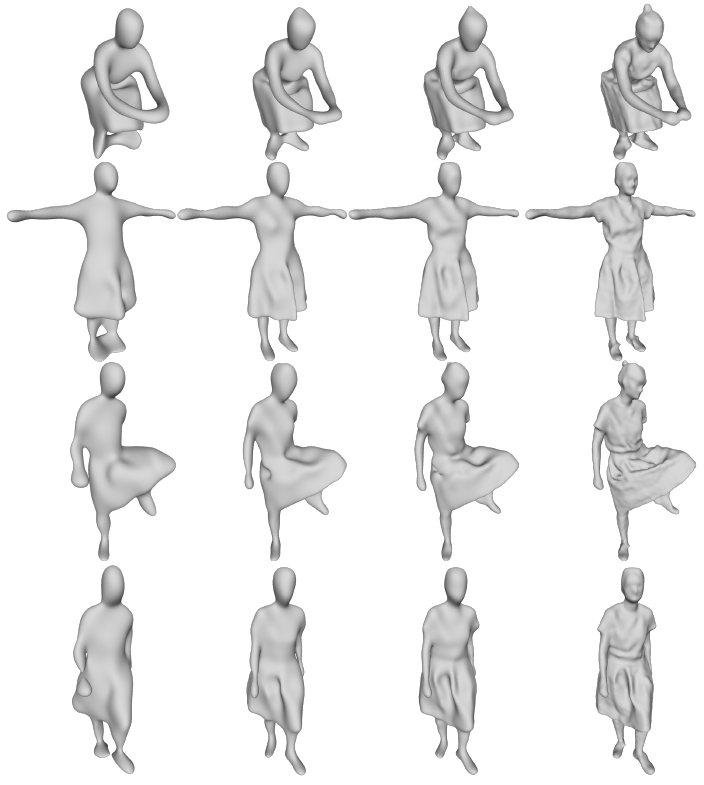}
    \end{overpic}
    %
    \caption{\textbf{Human Body Reconstructions} 
    with varying levels of detail, at resolutions $i=1,2,4$ and $12$.
    }
    \label{fig:people}
\end{figure*}


\begin{figure*}[t]
    \centering
    \begin{overpic}[width=0.97\textwidth]{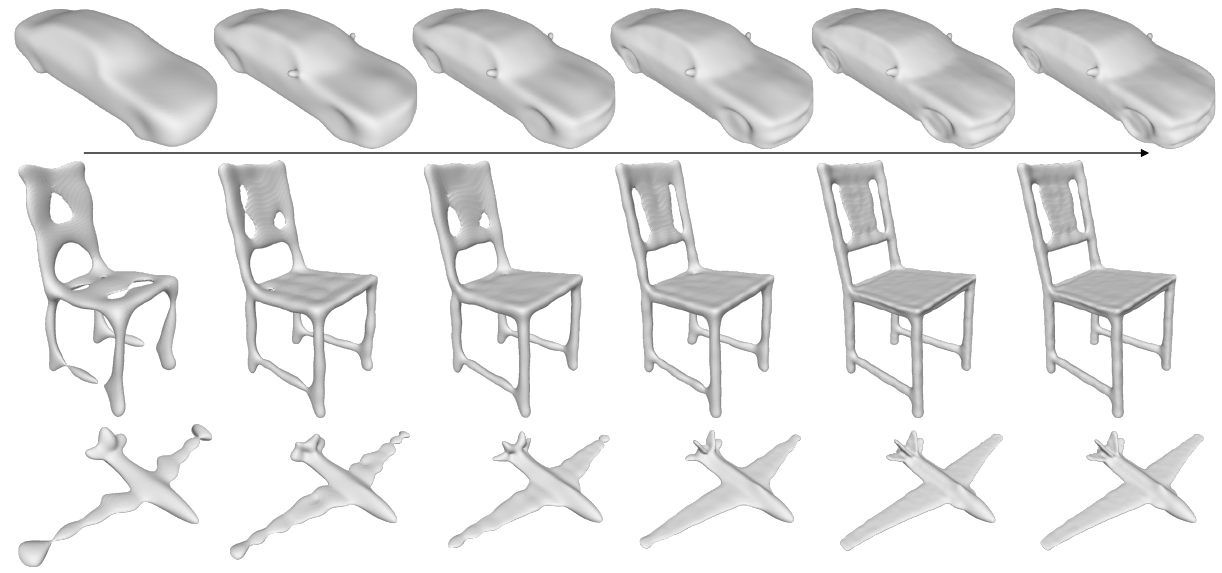}
        \put(3,35){$i=1$}
        \put(19,35){$i=2$}
        \put(36,35){$i=3$}
        \put(52,35){$i=5$}
        \put(68,35){$i=10$}
        \put(85,35){$i=12$}
    \end{overpic}
    %
    \caption{\textbf{Test Shape Reconstructions} 
    with varying levels of detail.
    %
    %
    The lowest possible level of detail ($i=1$) is smooth and already captures a very coarse structure of the shape.
    As $i$ increases, more details appear (up to a point of saturation).
    }
    \label{fig:example_supp}
\end{figure*}

\bg{
In \Cref{fig:people} we show qualitative results when training our network on a dataset of human body meshes. \Cref{fig:example_supp} displays additional results on the chairs, cars and airplanes categories. In both cases, our pipeline accurately captures details at high resolution, while also yielding smooth surfaces at low resolutions.
}


\section{Additional Comparisons with the Baselines}


\begin{table*}[]
  \small
  \centering
  \tabcolsep=0.05cm
  \begin{tabular}{r|cccccccccccc}
    \multicolumn{13}{l}{ } \\
    \multicolumn{13}{l}{\textbf{Metric: Chamfer Distance (CD, $\downarrow$)}} \\
   Layer \# : & 1 & 2 & 3 & 4 & 5 & 6 & 7 & 8 & 9 & 10 & 11 & 12 \\
   \hline
   \multicolumn{1}{l|}{\cellcolor[HTML]{C0C0C0} }    & \multicolumn{12}{|l}{\cellcolor[HTML]{C0C0C0}Cars}  \\
   {\begin{tabular}[c]{@{}r@{}}\baselineSin{} \vspace{-1.7mm} \\ \textit{\scriptsize{(1 / column)}} \vspace{-1.2mm} \end{tabular}} &
   8.0 &  7.9 &  7.1 &  6.8 &  \textbf{6.4} &  6.5 &  10.3 & 10.5 &  10.8 &  6.6 & 8.8 & 10.8 \\
   {\begin{tabular}[c]{@{}r@{}}\baselineRelu{} \vspace{-1.7mm} \\ \textit{\scriptsize{(1 / column)}} \vspace{-1.2mm} \end{tabular}} &
   \textbf{7.6} & \textbf{7.7} & \textbf{6.3} &  \textbf{6.0} &  8.6 & \textbf{6.0} &  \textbf{6.3} & \textbf{6.3} & \textbf{6.0} &  \textbf{6.0} & \textbf{6.1} &  \textbf{6.1} \\
   \begin{tabular}[c]{@{}r@{}}Ours \vspace{-1.7mm} \\ \textit{\scriptsize{(single net.)}} \vspace{-0.6mm} \end{tabular} &
   11.0 & 8.1  &  7.4  & 7.0  &  6.7  &  6.6  &  \textbf{6.3}  &  \textbf{6.3}  &  6.3  &  6.3  &  6.3  &  6.3  \\

   \multicolumn{1}{l|}{\cellcolor[HTML]{C0C0C0} }    & \multicolumn{12}{|l}{\cellcolor[HTML]{C0C0C0}Chairs}  \\
   {\begin{tabular}[c]{@{}r@{}}\baselineSin{} \vspace{-1.7mm} \\ \textit{\scriptsize{(1 / column)}} \vspace{-1.2mm} \end{tabular}}  &
   14.9 & 14.9 & 13.9 & 13.4 & 13.2 & 12.9 & 16.8 & 10.7 & 11.9 & 10.5 & 12.4 & 12.7 \\
   {\begin{tabular}[c]{@{}r@{}}\baselineRelu{} \vspace{-1.7mm} \\ \textit{\scriptsize{(1 / column)}} \vspace{-1.2mm} \end{tabular}}  &
   \textbf{13.1} & \textbf{10.0} &  \textbf{9.4}  & \textbf{9.1}  &  10.5 &  9.9  &  \textbf{8.5}  &  8.8  & 8.5  & 9.5  &  8.8  & 9.3   \\
   \begin{tabular}[c]{@{}r@{}}Ours \vspace{-1.7mm} \\ \textit{\scriptsize{(single net.)}} \vspace{-0.6mm} \end{tabular} &
   25.3 & 14.3 &  11.3 & 10.1 & \textbf{9.2}  & \textbf{8.6}  & \textbf{8.5}  & \textbf{8.3}  & \textbf{8.3}  &  \textbf{8.3}  &  \textbf{8.3}  &  \textbf{8.3} \\

   \multicolumn{1}{l|}{\cellcolor[HTML]{C0C0C0} }    & \multicolumn{12}{|l}{\cellcolor[HTML]{C0C0C0}Airplanes}  \\
   {\begin{tabular}[c]{@{}r@{}}\baselineSin{} \vspace{-1.7mm} \\ \textit{\scriptsize{(1 / column)}} \vspace{-1.2mm} \end{tabular}}  &
   \textbf{5.4} &  4.5 &  3.9 &  3.6 &  3.1 &  4.3 &  3.3 &  2.9 &  3.2 &  3.0 &  4.5 &  5.2 \\
   {\begin{tabular}[c]{@{}r@{}}\baselineRelu{} \vspace{-1.7mm} \\ \textit{\scriptsize{(1 / column)}} \vspace{-1.2mm} \end{tabular}}  &
   30.5 &  \textbf{3.0}  & \textbf{3.0}  & \textbf{2.7}  & \textbf{2.7}  &  \textbf{2.8}  &  \textbf{2.7}  & \textbf{2.6}  & 3.1 & \textbf{2.5} & \textbf{2.5} & \textbf{2.4}  \\
   \begin{tabular}[c]{@{}r@{}}Ours \vspace{-1.7mm} \\ \textit{\scriptsize{(single net.)}} \vspace{-0.6mm} \end{tabular} &
   12.2 & 5.8 & 4.4  & 3.8  & 3.2  & 2.9  & 2.8  & 2.7  & \textbf{2.7}  & 2.7 & 2.7  & 2.7  \\

   \multicolumn{13}{l}{ } \\
   \multicolumn{13}{l}{\textbf{Metric: Earth mover's Distance (ED, $\downarrow$)}} \\
    Layer \# : & 1 & 2 & 3 & 4 & 5 & 6 & 7 & 8 & 9 & 10 & 11 & 12 \\
    \hline
    \multicolumn{1}{l|}{\cellcolor[HTML]{C0C0C0} }    & \multicolumn{12}{|l}{\cellcolor[HTML]{C0C0C0}Cars}  \\
    {\begin{tabular}[c]{@{}r@{}}\baselineSin{} \vspace{-1.7mm} \\ \textit{\scriptsize{(1 / column)}} \vspace{-1.2mm} \end{tabular}} &
    4.0 &  3.9 &  3.6 &  3.5 &  3.4 &  3.3 &  3.5 &  3.5 &  3.5 &  3.3 &  3.5 &  3.5 \\
    {\begin{tabular}[c]{@{}r@{}}\baselineRelu{} \vspace{-1.7mm} \\ \textit{\scriptsize{(1 / column)}} \vspace{-1.2mm} \end{tabular}} &
    \textbf{3.2} &  \textbf{2.9} &  \textbf{2.8} &  \textbf{2.8} & \textbf{2.8} & \textbf{2.7} &  \textbf{2.7} &  \textbf{2.7} &  \textbf{2.7} &  \textbf{2.7} &  \textbf{2.7} & \textbf{2.7} \\
    \begin{tabular}[c]{@{}r@{}}Ours \vspace{-1.7mm} \\ \textit{\scriptsize{(single net.)}} \vspace{-0.6mm} \end{tabular} &
    3.8 & 3.2 &  3.0 &  2.9 &  \textbf{2.8} &  2.8 &  2.8 & 2.8 &  2.8 & 2.8 & 2.8 & 2.8 \\

    \multicolumn{1}{l|}{\cellcolor[HTML]{C0C0C0} }    & \multicolumn{12}{|l}{\cellcolor[HTML]{C0C0C0}Chairs}  \\
    {\begin{tabular}[c]{@{}r@{}}\baselineSin{} \vspace{-1.7mm} \\ \textit{\scriptsize{(1 / column)}} \vspace{-1.2mm} \end{tabular}}  &
    4.8 &  5.0 &  4.8 &  4.6 &  4.6 &  4.3 &  4.8 &  4.9 &  4.7 &  4.8 &  5.7 &  6.4 \\
    {\begin{tabular}[c]{@{}r@{}}\baselineRelu{} \vspace{-1.7mm} \\ \textit{\scriptsize{(1 / column)}} \vspace{-1.2mm} \end{tabular}}  &
    \textbf{4.5} & \textbf{3.9} & \textbf{3.8} & \textbf{3.7} & 3.8 & 3.8 & 3.5 & 3.6 & \textbf{3.4} & 3.7 & 3.5 &  3.6 \\
    \begin{tabular}[c]{@{}r@{}}Ours \vspace{-1.7mm} \\ \textit{\scriptsize{(single net.)}} \vspace{-0.6mm} \end{tabular} &
    7.0 &  5.4 &  4.4 & 4.0 & \textbf{3.7} & \textbf{3.5} & \textbf{3.4} & \textbf{3.4} & \textbf{3.4} &  \textbf{3.4} &  \textbf{3.4} & \textbf{3.4} \\

    \multicolumn{1}{l|}{\cellcolor[HTML]{C0C0C0} }    & \multicolumn{12}{|l}{\cellcolor[HTML]{C0C0C0}Airplanes}  \\
    {\begin{tabular}[c]{@{}r@{}}\baselineSin{} \vspace{-1.7mm} \\ \textit{\scriptsize{(1 / column)}} \vspace{-1.2mm} \end{tabular}}  &
    \textbf{3.3} &  3.3 &  2.7 &  2.5 &  2.1 &  2.2 &  2.3 &  2.1 &  2.1 &  1.8 &  15.7 & 16.1 \\
    {\begin{tabular}[c]{@{}r@{}}\baselineRelu{} \vspace{-1.7mm} \\ \textit{\scriptsize{(1 / column)}} \vspace{-1.2mm} \end{tabular}}  &
    \textbf{3.3} & \textbf{2.0} &  \textbf{1.9} &  \textbf{2.0} &  \textbf{1.6} &  1.8 &  \textbf{1.6} &  1.8 &  1.9 &  \textbf{1.6} &  1.6 & 1.6 \\
    \begin{tabular}[c]{@{}r@{}}Ours \vspace{-1.7mm} \\ \textit{\scriptsize{(single net.)}} \vspace{-0.6mm} \end{tabular} &
    4.3 &  3.0 &  2.7 &  2.3 &  1.9 &  \textbf{1.7} & \textbf{1.6} &  \textbf{1.5} &  \textbf{1.5} & \textbf{1.6} & \textbf{1.5} &  \textbf{1.5} \\

  \multicolumn{13}{l}{ } \\
  \multicolumn{13}{l}{\textbf{Metric: Surface Regularity (SR, $\downarrow$)}} \\
  Layer \# : & 1 & 2 & 3 & 4 & 5 & 6 & 7 & 8 & 9 & 10 & 11 & 12 \\
  \hline
  \multicolumn{1}{l|}{\cellcolor[HTML]{C0C0C0} }    & \multicolumn{12}{|l}{\cellcolor[HTML]{C0C0C0}Cars}  \\
  {\begin{tabular}[c]{@{}r@{}}\baselineSin{} \vspace{-1.7mm} \\ \textit{\scriptsize{(1 / column)}} \vspace{-1.2mm} \end{tabular}} &
  9.5  & 9.5  & 9.8  & 9.9  & 9.8  & 9.5  & 9.7  & 9.7 & 9.6 & 10.4 & 10.1 & 9.6 \\
  {\begin{tabular}[c]{@{}r@{}}\baselineRelu{} \vspace{-1.7mm} \\ \textit{\scriptsize{(1 / column)}} \vspace{-1.2mm} \end{tabular}} &
  11.7 & 11.2 & 11.1 & 10.9 & 10.8 & 11.1 & 10.7 & 10.9 & 11.0 & 11.0 & 11.1 & 11.4 \\
  \begin{tabular}[c]{@{}r@{}}Ours \vspace{-1.7mm} \\ \textit{\scriptsize{(single net.)}} \vspace{-0.6mm} \end{tabular} &
  \textbf{4.9} & \textbf{5.5} & \textbf{5.9} & \textbf{6.1} & \textbf{6.4} & \textbf{6.6} & \textbf{6.9} & \textbf{7.1} & \textbf{7.2} & \textbf{7.3} & \textbf{7.3} & \textbf{7.3} \\

  \multicolumn{1}{l|}{\cellcolor[HTML]{C0C0C0} }    & \multicolumn{12}{|l}{\cellcolor[HTML]{C0C0C0}Chairs}  \\
  {\begin{tabular}[c]{@{}r@{}}\baselineSin{} \vspace{-1.7mm} \\ \textit{\scriptsize{(1 / column)}} \vspace{-1.2mm} \end{tabular}}  &
  13.2 & 13.0 & 12.2 & 12.0 & 12.6 & 13.2 & 14.2 & 16.0 & 14.4 & 13.9 & 18.8 & 19.0 \\
  {\begin{tabular}[c]{@{}r@{}}\baselineRelu{} \vspace{-1.7mm} \\ \textit{\scriptsize{(1 / column)}} \vspace{-1.2mm} \end{tabular}}  &
  14.7 & 13.0 & 12.6 & 12.4 & 12.6 & 12.3 & 12.5 & 12.5 & 12.8 & 13.1 & 13.1 & 13.2 \\
  \begin{tabular}[c]{@{}r@{}}Ours \vspace{-1.7mm} \\ \textit{\scriptsize{(single net.)}} \vspace{-0.6mm} \end{tabular} &
  \textbf{6.8} & \textbf{7.8} & \textbf{7.7} & \textbf{7.8} & \textbf{7.9} & \textbf{8.1} & \textbf{8.2} & \textbf{8.3 }& \textbf{8.4} & \textbf{8.4} & \textbf{8.4} & \textbf{8.4} \\

  \multicolumn{1}{l|}{\cellcolor[HTML]{C0C0C0} }    & \multicolumn{12}{|l}{\cellcolor[HTML]{C0C0C0}Airplanes}  \\
  {\begin{tabular}[c]{@{}r@{}}\baselineSin{} \vspace{-1.7mm} \\ \textit{\scriptsize{(1 / column)}} \vspace{-1.2mm} \end{tabular}}  &
  15.9 & 14.0 & 12.9 & 13.7 & 13.6 & \textbf{13.7} & \textbf{13.6} & 14.7 & \textbf{14.1} & 14.8 & 15.2 & 16.1 \\
  {\begin{tabular}[c]{@{}r@{}}\baselineRelu{} \vspace{-1.7mm} \\ \textit{\scriptsize{(1 / column)}} \vspace{-1.2mm} \end{tabular}}  &
  15.3 & 15.0 & 14.6 & 14.5 & 15.2 & 14.6 & 14.9 & 15.5 & 14.4 & 16.2 & 16.5 & 14.8 \\
  \begin{tabular}[c]{@{}r@{}}Ours \vspace{-1.7mm} \\ \textit{\scriptsize{(single net.)}} \vspace{-0.6mm} \end{tabular} &
  \textbf{9.6} & \textbf{12.2} & \textbf{12.8} & \textbf{13.0} & \textbf{13.4} & \textbf{13.7} & 13.8 & \textbf{14.1} & 14.4 & \textbf{14.4} & \textbf{14.5} & \textbf{14.5} \\
 \end{tabular}
 \vspace{2mm}
 \caption{\textbf{Unseen test shapes}: Chamfer Distances (CD), Earth mover's Distances (ED) and Surface Regularity (SR), for all depths. Lower values indicate better results for all metrics. We achieve comparable best accuracy (CD, ED) while being consistently smoother (SR) than the best baseline (\baselineRelu{}).}
  \label{tab:supp_full_testset}
\end{table*}

\begin{table*}[]
  \small
  \centering
  \tabcolsep=0.05cm
  \begin{tabular}{r|cccccccccccc}
    \multicolumn{13}{l}{ } \\
    \multicolumn{13}{l}{\textbf{Metric: Chamfer Distance (CD, $\downarrow$)}} \\
   Layer \# : & 1 & 2 & 3 & 4 & 5 & 6 & 7 & 8 & 9 & 10 & 11 & 12 \\
   \hline
   \multicolumn{1}{l|}{\cellcolor[HTML]{C0C0C0} }    & \multicolumn{12}{|l}{\cellcolor[HTML]{C0C0C0}Cars}  \\
   {\begin{tabular}[c]{@{}r@{}}\baselineSin{} \vspace{-1.7mm} \\ \textit{\scriptsize{(1 / column)}} \vspace{-1.2mm} \end{tabular}} &
   10.7 & 10.7 & 9.8 & 9.3 & 8.2 & 7.9 & 7.6 & 7.5 & 7.3 & 8.4 & 10.6 & 10.7 \\
   {\begin{tabular}[c]{@{}r@{}}\baselineRelu{} \vspace{-1.7mm} \\ \textit{\scriptsize{(1 / column)}} \vspace{-1.2mm} \end{tabular}} &
   \textbf{7.8} & \textbf{6.1} & \textbf{5.8} & \textbf{5.7} & \textbf{5.7} & \textbf{5.6} & \textbf{5.7} & \textbf{5.7} & 5.7 & 5.7 & 5.7 & 5.7 \\
   \begin{tabular}[c]{@{}r@{}}Ours \vspace{-1.7mm} \\ \textit{\scriptsize{(single net.)}} \vspace{-0.6mm} \end{tabular} &
   10.4 & 7.5 & 6.7 & 6.4 & 6.0 & 5.9 & \textbf{5.7} & \textbf{5.7} & \textbf{5.6} & \textbf{5.6} & \textbf{5.6} & \textbf{5.6} \\

   \multicolumn{1}{l|}{\cellcolor[HTML]{C0C0C0} }    & \multicolumn{12}{|l}{\cellcolor[HTML]{C0C0C0}Chairs}  \\
   {\begin{tabular}[c]{@{}r@{}}\baselineSin{} \vspace{-1.7mm} \\ \textit{\scriptsize{(1 / column)}} \vspace{-1.2mm} \end{tabular}}  &
   29.0 & 27.5 & 13.1 & 12.2 & 9.1 & 7.9 & 7.1 & 6.2 & 6.6 & 6.0 & 11.9 & 14.9 \\
   {\begin{tabular}[c]{@{}r@{}}\baselineRelu{} \vspace{-1.7mm} \\ \textit{\scriptsize{(1 / column)}} \vspace{-1.2mm} \end{tabular}}  &
   \textbf{11.1} & \textbf{7.9} & \textbf{6.8} & \textbf{6.4} & \textbf{6.2} & \textbf{6.0} & 5.9 & 6.0 & 5.8 & 5.8 & 5.9 & 6.1 \\
   \begin{tabular}[c]{@{}r@{}}Ours \vspace{-1.7mm} \\ \textit{\scriptsize{(single net.)}} \vspace{-0.6mm} \end{tabular} &
   17.8 & 9.1 & 7.8 & 7.0 & 6.4 & \textbf{6.0} & \textbf{5.8} & \textbf{5.7} & \textbf{5.6} & \textbf{5.6} & \textbf{5.6} & \textbf{5.6} \\

   \multicolumn{1}{l|}{\cellcolor[HTML]{C0C0C0} }    & \multicolumn{12}{|l}{\cellcolor[HTML]{C0C0C0}Airplanes}  \\
   {\begin{tabular}[c]{@{}r@{}}\baselineSin{} \vspace{-1.7mm} \\ \textit{\scriptsize{(1 / column)}} \vspace{-1.2mm} \end{tabular}}  &
   21.3 & 21.5 & 7.5 & 6.1 & 6.2 & 5.8 & 6.2 & 6.0 & 6.0 & 5.4 & 7.6 & 9.1 \\
   {\begin{tabular}[c]{@{}r@{}}\baselineRelu{} \vspace{-1.7mm} \\ \textit{\scriptsize{(1 / column)}} \vspace{-1.2mm} \end{tabular}}  &
   \textbf{7.7} & \textbf{2.7} & \textbf{2.3} & \textbf{2.3} & \textbf{2.2} & 2.4 & 2.3 & 2.3 & 2.2 & 2.2 & 2.4 & 2.1 \\
   \begin{tabular}[c]{@{}r@{}}Ours \vspace{-1.7mm} \\ \textit{\scriptsize{(single net.)}} \vspace{-0.6mm} \end{tabular} &
   15.0 & 4.0 & 3.0 & 2.5 & \textbf{2.2} & \textbf{2.1} & \textbf{2.0} & \textbf{1.9} & \textbf{1.8} & \textbf{1.8} & \textbf{1.8} & \textbf{1.8}\\

   \multicolumn{13}{l}{ } \\
   \multicolumn{13}{l}{\textbf{Metric: Earth mover's Distance (ED, $\downarrow$)}} \\
    Layer \# : & 1 & 2 & 3 & 4 & 5 & 6 & 7 & 8 & 9 & 10 & 11 & 12 \\
    \hline
    \multicolumn{1}{l|}{\cellcolor[HTML]{C0C0C0} }    & \multicolumn{12}{|l}{\cellcolor[HTML]{C0C0C0}Cars}  \\
    {\begin{tabular}[c]{@{}r@{}}\baselineSin{} \vspace{-1.7mm} \\ \textit{\scriptsize{(1 / column)}} \vspace{-1.2mm} \end{tabular}} &
    4.2 & 4.2 & 3.9 & 3.8 & 3.6 & 3.5 & 3.4 & 3.4 & 3.4 & 3.6 & 4.0 & 3.6 \\
    {\begin{tabular}[c]{@{}r@{}}\baselineRelu{} \vspace{-1.7mm} \\ \textit{\scriptsize{(1 / column)}} \vspace{-1.2mm} \end{tabular}} &
    \textbf{3.1} & \textbf{2.7} & \textbf{2.7} & \textbf{2.6} & \textbf{2.6} & \textbf{2.6} & \textbf{2.6} & \textbf{2.6} & \textbf{2.6} & \textbf{2.6} & \textbf{2.6} & \textbf{2.6} \\
    \begin{tabular}[c]{@{}r@{}}Ours \vspace{-1.7mm} \\ \textit{\scriptsize{(single net.)}} \vspace{-0.6mm} \end{tabular} &
    3.6 & 3.0 & 2.9 & 2.8 & 2.7 & \textbf{2.6} & \textbf{2.6} & \textbf{2.6} & \textbf{2.6} & \textbf{2.6} & \textbf{2.6} & \textbf{2.6} \\

    \multicolumn{1}{l|}{\cellcolor[HTML]{C0C0C0} }    & \multicolumn{12}{|l}{\cellcolor[HTML]{C0C0C0}Chairs}  \\
    {\begin{tabular}[c]{@{}r@{}}\baselineSin{} \vspace{-1.7mm} \\ \textit{\scriptsize{(1 / column)}} \vspace{-1.2mm} \end{tabular}}  &
    14.7 & 14.0 & 7.5 & 6.5 & 5.1 & 4.4 & 4.1 & 3.6 & 3.3 & 3.0 & 6.9 & 8.0 \\
    {\begin{tabular}[c]{@{}r@{}}\baselineRelu{} \vspace{-1.7mm} \\ \textit{\scriptsize{(1 / column)}} \vspace{-1.2mm} \end{tabular}}  &
    \textbf{4.2} & \textbf{3.5} & \textbf{3.1} & \textbf{2.9} & \textbf{2.9} & \textbf{2.8} & \textbf{2.7} & 2.8 & \textbf{2.7} & \textbf{2.7} & \textbf{2.7} & 2.8 \\
    \begin{tabular}[c]{@{}r@{}}Ours \vspace{-1.7mm} \\ \textit{\scriptsize{(single net.)}} \vspace{-0.6mm} \end{tabular} &
    5.6 & 3.7 & 3.3 & 3.1 & 3.0 & \textbf{2.8} & 2.8 & \textbf{2.7} & \textbf{2.7} & \textbf{2.7} & \textbf{2.7} & \textbf{2.7}\\

    \multicolumn{1}{l|}{\cellcolor[HTML]{C0C0C0} }    & \multicolumn{12}{|l}{\cellcolor[HTML]{C0C0C0}Airplanes}  \\
    {\begin{tabular}[c]{@{}r@{}}\baselineSin{} \vspace{-1.7mm} \\ \textit{\scriptsize{(1 / column)}} \vspace{-1.2mm} \end{tabular}}  &
    4.2 & 4.3 & 2.1 & 1.7 & 1.6 & 1.5 & 1.5 & 1.5 & 1.5 & 1.4 & 2.2 & 2.4 \\
    {\begin{tabular}[c]{@{}r@{}}\baselineRelu{} \vspace{-1.7mm} \\ \textit{\scriptsize{(1 / column)}} \vspace{-1.2mm} \end{tabular}}  &
    \textbf{2.2} & \textbf{1.5} & \textbf{1.4} & \textbf{1.4} & \textbf{1.3} & 1.3 & 1.3 & 1.3 & 1.3 & 1.3 & 1.3 & 1.3 \\
    \begin{tabular}[c]{@{}r@{}}Ours \vspace{-1.7mm} \\ \textit{\scriptsize{(single net.)}} \vspace{-0.6mm} \end{tabular} &
    4.4 & 2.1 & 1.7 & 1.5 & \textbf{1.3} & \textbf{1.2} & \textbf{1.1} & \textbf{1.0} & \textbf{1.0} & \textbf{1.0} & \textbf{1.0} & \textbf{1.0} \\

  \multicolumn{13}{l}{ } \\
  \multicolumn{13}{l}{\textbf{Metric: Surface Regularity (SR, $\downarrow$)}} \\
  Layer \# : & 1 & 2 & 3 & 4 & 5 & 6 & 7 & 8 & 9 & 10 & 11 & 12 \\
  \hline
  \multicolumn{1}{l|}{\cellcolor[HTML]{C0C0C0} }    & \multicolumn{12}{|l}{\cellcolor[HTML]{C0C0C0}Cars}  \\
  {\begin{tabular}[c]{@{}r@{}}\baselineSin{} \vspace{-1.7mm} \\ \textit{\scriptsize{(1 / column)}} \vspace{-1.2mm} \end{tabular}} &
  6.8 & 6.8 & 7.4 & 7.7 & 8.1 & 8.5 & 8.8 & 8.8 & 9.0 & 8.4 & 13.8 & 14.8 \\
  {\begin{tabular}[c]{@{}r@{}}\baselineRelu{} \vspace{-1.7mm} \\ \textit{\scriptsize{(1 / column)}} \vspace{-1.2mm} \end{tabular}} &
  9.7 & 9.3 & 9.3 & 9.2 & 9.5 & 9.5 & 9.6 & 9.2 & 9.5 & 9.6 & 9.6 & 9.5 \\
  \begin{tabular}[c]{@{}r@{}}Ours \vspace{-1.7mm} \\ \textit{\scriptsize{(single net.)}} \vspace{-0.6mm} \end{tabular} &
  \textbf{4.9} & \textbf{5.6} & \textbf{6.0} & \textbf{6.3} & \textbf{6.7} & \textbf{7.0} & \textbf{7.3} & \textbf{7.6} & \textbf{7.7} & \textbf{7.7} & \textbf{7.8} & \textbf{7.8} \\

  \multicolumn{1}{l|}{\cellcolor[HTML]{C0C0C0} }    & \multicolumn{12}{|l}{\cellcolor[HTML]{C0C0C0}Chairs}  \\
  {\begin{tabular}[c]{@{}r@{}}\baselineSin{} \vspace{-1.7mm} \\ \textit{\scriptsize{(1 / column)}} \vspace{-1.2mm} \end{tabular}}  &
  9.6 & 9.5 & 9.0 & 9.0 & 9.4 & 9.8 & 10.3 & 11.3 & 10.8 & 10.9 & 15.7 & 15.9 \\
  {\begin{tabular}[c]{@{}r@{}}\baselineRelu{} \vspace{-1.7mm} \\ \textit{\scriptsize{(1 / column)}} \vspace{-1.2mm} \end{tabular}}  &
  12.6 & 10.9 & 10.5 & 10.4 & 10.6 & 10.4 & 10.6 & 10.5 & 10.8 & 11.0 & 11.1 & 10.9 \\
  \begin{tabular}[c]{@{}r@{}}Ours \vspace{-1.7mm} \\ \textit{\scriptsize{(single net.)}} \vspace{-0.6mm} \end{tabular} &
  \textbf{6.8} & \textbf{7.6} & \textbf{7.7} & \textbf{7.9} & \textbf{8.1} & \textbf{8.3} & \textbf{8.4} & \textbf{8.6} & \textbf{8.7} & \textbf{8.8} & \textbf{8.8} & \textbf{8.8} \\

  \multicolumn{1}{l|}{\cellcolor[HTML]{C0C0C0} }    & \multicolumn{12}{|l}{\cellcolor[HTML]{C0C0C0}Airplanes}  \\
  {\begin{tabular}[c]{@{}r@{}}\baselineSin{} \vspace{-1.7mm} \\ \textit{\scriptsize{(1 / column)}} \vspace{-1.2mm} \end{tabular}}  &
  13.8 & 13.8 & 14.1 & 14.4 & 14.7 & 14.9 & 15.2 & 15.4 & 15.6 & 15.8 & 17.8 & 19.3 \\
  {\begin{tabular}[c]{@{}r@{}}\baselineRelu{} \vspace{-1.7mm} \\ \textit{\scriptsize{(1 / column)}} \vspace{-1.2mm} \end{tabular}}  &
  17.4 & 16.6 & 16.5 & 16.6 & 16.7 & 16.7 & 16.9 & 16.8 & 16.8 & 17.0 & 16.9 & 17.1 \\
  \begin{tabular}[c]{@{}r@{}}Ours \vspace{-1.7mm} \\ \textit{\scriptsize{(single net.)}} \vspace{-0.6mm} \end{tabular} &
  \textbf{9.7} & \textbf{12.2} & \textbf{12.8} & \textbf{13.0} & \textbf{13.3} & \textbf{13.6} & \textbf{13.8} & \textbf{14.2} & \textbf{14.5} & \textbf{14.6} & \textbf{14.6} & \textbf{14.6} \\
 \end{tabular}
 \vspace{2mm}
 \caption{\textbf{Training shapes}: Chamfer Distances (CD), Earth mover's Distances (ED) and Surface Regularity (SR), for all depths. Lower values indicate better results for all metrics. We achieve comparable best accuracy (CD, ED) while being consistently smoother (SR) than the best baseline (\baselineRelu{}).}
  \label{tab:supp_full_trainset}
\end{table*}

In \Cref{tab:supp_full_testset,tab:supp_full_trainset} we report metrics at all depths levels for our method and the two baselines \baselineRelu{} and \baselineSin{} on both unseen test shapes and training ones.
Their interpretation is consistent with the ones reported in the main text: our pipeline achieves comparable accuracy levels, while being consistently smoother.


\begin{figure*}
    \centering
    \includegraphics[width=0.8\textwidth]{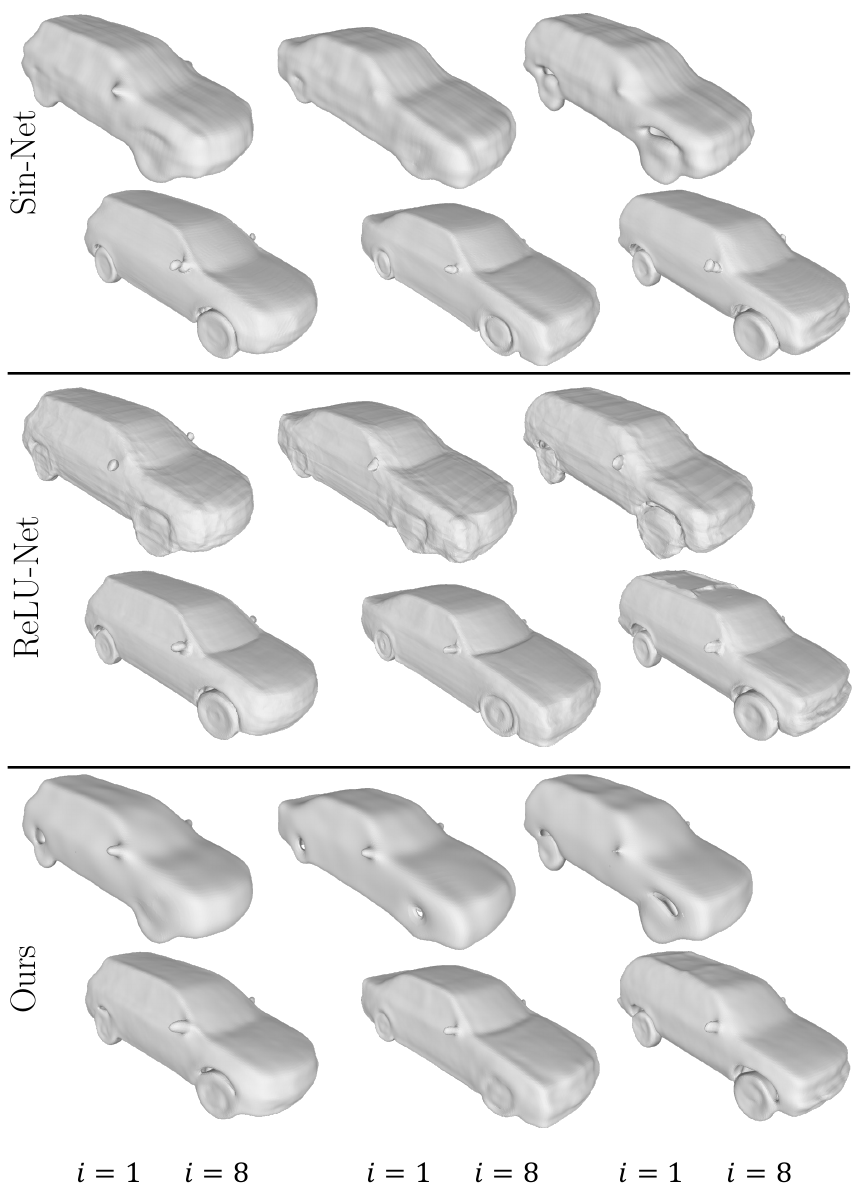}
    %
    \caption{\textbf{Unseen test cars} reconstructed by \baselineSin{} (\textbf{top}), \baselineRelu{} (\textbf{middle}) and our network (\textbf{bottom}), at levels of detail $i=1$ and $i=8$.
    }
    \label{fig:ex_cars_p}
\end{figure*}


\begin{figure*}
    \centering
    \includegraphics[width=0.8\textwidth]{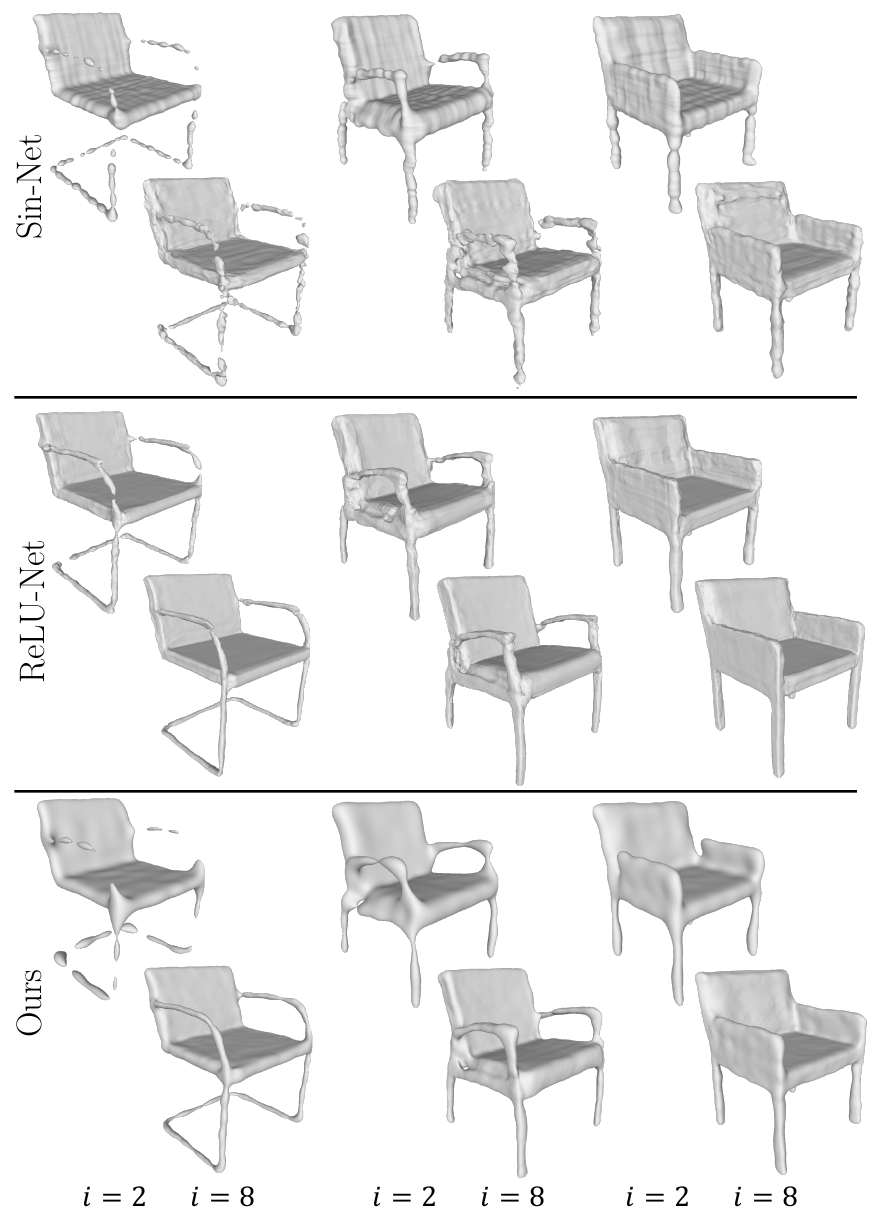}
    %
    \caption{\textbf{Unseen test chairs} reconstructed by \baselineSin{} (\textbf{top}), \baselineRelu{} (\textbf{middle}) and our network (\textbf{bottom}), at levels of detail $i=2$ and $i=8$.
    }
    \label{fig:ex_chairs_p}
\end{figure*}

In \Cref{fig:ex_cars_p,fig:ex_chairs_p} we show unseen cars and chairs reconstructed by \baselineSin{}, \baselineRelu{} and our network.

\end{document}